\pdfoutput=1

\documentclass[10pt,journal,compsoc]{IEEEtran}

\usepackage{textcomp}

\ifCLASSOPTIONcompsoc
  \usepackage[nocompress]{cite}
\else
  \usepackage{cite}
\fi

\ifCLASSINFOpdf
  \usepackage[pdftex]{graphicx}
  \graphicspath{{.}}
  \DeclareGraphicsExtensions{.pdf,.jpeg,.png}
\else
\fi

\def\eg{\emph{e.g.\/}}
\def\ie{\emph{i.e.\/}}
\def\etc{\emph{etc.\/}}
\def\etal{\emph{et al.\/}}
\def\vs{\emph{vs.\/}}

\begin{document}

\title{Recognizing Detailed Human Context In-the-Wild from Smartphones and Smartwatches}

\author{Yonatan~Vaizman$^{\ast\dagger}$
        Katherine~Ellis$^{\ast}$
        and~Gert~Lanckriet,~\IEEEmembership{senior~member,~IEEE}
\IEEEcompsocitemizethanks{\IEEEcompsocthanksitem{Y. Vaizman, K. Ellis and G. Lanckriet were with the Department
of Electrical and Computer Engineering, University of California, San Diego.}\protect
\IEEEcompsocthanksitem{$\ast$ Y. Vaizman and K. Ellis are equal contributors.}
\IEEEcompsocthanksitem{$\dagger$ Corresponding author. Email: yvaizman@eng.ucsd.edu}}\\
This version was accepted to IEEE Pervasive Computing magazine.\\
Please cite the official (edited and stylized) paper, which appears in IEEE Pervasive Computing, vol. 16, no. 4, October-December 2017, pp. 62-74.}


\IEEEtitleabstractindextext{%
\begin{abstract}
The ability to automatically recognize a person's behavioral context can contribute to health monitoring, aging care and many other domains. Validating context recognition \emph{in-the-wild} is crucial to promote practical applications that work in real-life settings. We collected over 300k minutes of sensor data with context labels from 60 subjects. Unlike previous studies, our subjects used their own personal phone, in any way that was convenient to them, and engaged in their routine in their natural environments. Unscripted behavior and unconstrained phone usage resulted in situations that are harder to recognize. We demonstrate how fusion of multi-modal sensors is important for resolving such cases.
We present a baseline system, and encourage researchers to use our public dataset to compare methods and improve context recognition in-the-wild.
\end{abstract}

\begin{IEEEkeywords}
Context awareness, Mobile sensors, Artificial intelligence, Machine learning, Human activity recognition.
\end{IEEEkeywords}}

\maketitle

\IEEEdisplaynontitleabstractindextext

%
\IEEEpeerreviewmaketitle

\IEEEraisesectionheading{\section{Introduction}\label{sec:introduction}}
The ability to automatically recognize a person's context (\ie,~where they are, what they are doing, who they are with,~\etc) is greatly beneficial in many domains. Health monitoring and lifestyle interventions have traditionally been based on manual, subjective reporting~\cite{servick2015mind}, sometimes by end-of-day recalling~\cite{basner2007american}. These can improve with automatic (frequent, effortless, and objective) detection of behaviors like exercise, eating, sleeping, or mental states like stress. Just-in-time interventions (\eg~for addiction) often prompt the patient at arbitrary times of the day, possibly missing times when the patient needs support the most~\cite{nahum2014just}. Automatically recognizing context will help detect critical times and offer immediate support (\eg~an alcoholic patient may be in high risk of craving or lapse when the context is ``at a bar, with friends'').

The biomedical research community acknowledges the effects of behavior, lifestyle and environment on health, disease and treatments~\cite{intille2016precision}. Automatic context recognition tools will be essential to incorporate behavioral and exposure aspects into large scale studies and to tailor appropriate treatment for patients. The range of measured exposures should be broad and cover diverse life style and environmental aspects. Commercial tools that offer superficial recognition (\eg~of walking, running, and driving) will not suffice.
Personal assistant systems can adjust to context and better serve the user.
Aging care programs can use automated logging of older adults' behavior to detect early signs of cognitive impairment, monitor functional independence, and support aging at home~\cite{lee2015sensor}.

In order for such applications to succeed in large scale, the context recognition component has to be unobtrusive and to work smoothly, without requiring the person to adjust their behavior. It is important that research emulates real-world settings, where such applications will eventually be deployed. In this work we promote context recognition \emph{in-the-wild}, meaning capturing people's authentic behavior in their natural environments, with the use of every-day devices --- smartphones and smartwatches. We address the difficulty that in-the-wild conditions add, and show how multi-modal sensors can help.

\subsection*{Related work\label{subsec:related_work}}
It is common for people to have their phone close to them most of the time~\cite{dey2011getting}. This growing trend, and the variety of built-in sensors, make phones popular agents for recognizing human behavior. Smartwatches are a useful sensing addition. While capturing informative signals about hand and arm motion, they remain very natural to wear and don't add any burden to the user.

Previous works have shown the advantage of fusing sensors of different modalities, from smartphones and smartwatches, to improve recognition of basic movement activities~\cite{guiry2014multi} and more complex activities, like smoking or drinking coffee~\cite{shoaib2015towards}. However, most past works have collected data under heavily controlled conditions, with researchers instructing subjects to perform scripted tasks. Fitting models to recognize prescribed activities may result in poor generalization to real life scenarios~\cite{kerr2016objective}.
To promote real-life working applications, we argue that research has to be done in natural and realistic settings, satisfying four \emph{in-the-wild} conditions:\\
\begin{enumerate}
	\item Naturally used devices. Introducing a foreign device to the user adds a burden and harms natural behavior. Ideally, subjects would use their own phone, and possibly additional convenient devices, like watches.
	\item Unconstrained device placement. It has been shown that the placement and orientation of sensors have a great influence on the success of recognition~\cite{kunze2014sensor}. However, this does not mean we should avoid this difficulty by forcing specific placement --- a practical real-world application cannot restrict users to keep their phone in pant pocket for the recognition to work. Instead, research should address the variability in device placement as a challenge, and provide solutions to overcome it.
	\item Natural environment. The recorded behavior should be in the subjects' natural environment and on their own free time. They should not be instructed where or when to perform their activity.
	\item Natural behavioral content. In many works the researchers instructed subjects to perform scripted tasks~\cite{guiry2014multi,shoaib2015towards}. The recorded behavior was then simulated, and not natural. Other works let the subjects behave on their own time, but still prescribed a list of targeted activities~\cite{ganti2010multisensor,pirsiavash2012detecting}, which may cause the subject to perform actions they are not used to, like ``vacuum cleaning''. In-the-wild studies should record the behavior that is natural to each individual subject.
\end{enumerate}
A major challenge is acquiring labels of the behavioral context. Attaining in-the-wild conditions usually trades off with other aspects of the data collection effort, resulting in fewer labeled examples, smaller range of interest labels or compromised privacy of the subjects.

Previous research addressed some aspects of in-the-wild data collection in different ways.
Han~\etal~\cite{han2012comprehensive} designed a decision-tree architecture that activates predetermined sensors to differentiate eight ambulatory and transportation states. Such a hand-crafted system is hard to scale to more contexts. They validated their system with an observer that followed a single user.
Ordonez~\etal~\cite{ordonez2013activity} installed a set of state-change sensors around a home to detect daily home activities. While such sensors are un-obtrusive and maintain natural behavior, the complicated device setup limits the deployment of data collection and practical applications. It also cannot track the person outside of the monitored environment.
Dong~\etal~\cite{dong2014detecting} targeted eating periods and used an unnatural setup of having a smartphone bound to the wrist. Subjects had to mark start times of eating, and after data collection they reviewed and corrected their markings. This resulted in 449 hours of data with 116 eating periods from 43 subjects.
Rahman~\etal~\cite{rahman2014towards} compared different approaches for subjects to self-report their stress level (immediately or by recalling later). They suggested a compromise approach where the subject can report on their own time but with the help of cues (like location or surrounding sound level) to remember how they felt at specific times of the day.

Choudhury~\etal~\cite{choudhury2008mobile} designed a system to address the requirements for a practical context recognition system, including unobtrusive lightweight devices, long battery life and multi-modal sensing. However, most of their validation was done on controlled data, collected in specific locations, with constrained positioning of device, and with a sequence of 8 activities that was scripted, observed, and repeated by 12 subjects. Consolvo~\etal~\cite{consolvo2008activity} utilized the same system (trained on the controlled data) in a field study of an application to promote physical activity. The mobile app used a combination of the automated recognition (of walking, cycling,~\etc) and manual editing of a daily journal.

Hemminki~\etal~\cite{hemminki2013accelerometer} targeted detecting transportation modes and specifically designed features that would be less sensitive to placement of the phone.
Ganti~\etal~\cite{ganti2010multisensor} gave eight subjects a Nokia N95 phone for a period of eight weeks and asked them to go about their regular routine and use the phone for recording whenever they can, in any location or time-of-day. The phone was constrained to be in the pocket or pouch. The interface allowed selecting an activity from a set of eight activities and marking when you start and when you finish. They collected a total of 80 hours.
Khan~\etal~\cite{khan2014activity} targeted recognition of 15 activities. To collect measurements and annotations, they handed a NEXUS phone to subjects for a month. Subjects were free to perform the prescribed activities on their own time and they used the phone to mark the beginning and end of the selected activity. They collected about 3000 examples per activity from 30 subjects, plus a follow-up validation with eight subjects using the trained real-time recognition system.

In Yatani~\etal's~\cite{yatani2012bodyscope} out-of-lab study, five subject wore a phone around their neck to take egocentric snapshots that were later used to label the activity.
A similar label acquisition strategy was taken in large scale by Ellis~\etal~\cite{ellis2014multi} with 40 subjects who recorded hip-mounted accelerometer and GPS data from routine behavior in natural environment for several days. The subjects wore a SenseCam device around their neck, which took snapshots periodically, and the thousands of images were later used by research assistants to annotate the activity.
Pirsavash~\etal\cite{pirsiavash2012detecting} used a GoPro video camera for both sensor measurements and ground truth labels. The subjects wore the device around the chest in a single morning at their own home, and were prescribed a list of home activities to perform with no extra specifications. They recorded over 10 hours of video from 12 people and later annotated household objects and activities for about 30k frames (every second). Their dataset is publicly available.
While the camera approach may generate more reliable labels in certain cases, the unnatural and uncomfortable equipment compromises natural behavior. Furthermore, offline annotation of images is costly, making it hard to scale, and violates the privacy of the subjects and people around them. The alternative of self reporting has the advantage of collecting labels when a camera is not present (\eg~``shower''), when the context is not clearly visible in the image (\eg~``singing''), or when the subject knows best what is happening (\eg~``with family'' \vs~``with friends'').

\subsection*{Our work\label{subsec:our_work}}
In this work we use smartphone and smartwatch sensors to recognize detailed situations of people in their natural behavior. We collected labeled data from over 300k minutes from 60 subjects. Every minute has multi-sensor measurements and is annotated with relevant context labels. To the best of our knowledge, this dataset, which is publicly available, is far larger in scale compared to others collected in the field.
Similar to \cite{dong2014detecting,ganti2010multisensor,khan2014activity}, we rely on self-report. Unlike those works, our data collection app offers an extensive menu of over 100 context labels and the ability to select combinations of relevant labels. This facilitates natural behavior from the subjects, \eg~they are free to ``run on a treadmill'', while ``watching TV'' if it is natural to them (in \cite{ganti2010multisensor,khan2014activity}, subjects were forced to choose one activity, possibly causing them to act unnaturally). This also provides rich descriptions of context, as combinations of different aspects, like environment, activities, company, body posture. Similar combinatorial representations were previously used to describe objects and actions in images~\cite{pirsiavash2012detecting} and locations, objects, humans, and animals in sound clips~\cite{rossi2012recognizing}. In those cases annotation was done offline, but in our case, attaining detailed labeling by self-reporting requires attention and effort from the subjects. To mitigate it, our app's interface offers many reporting-mechanisms to minimize interaction time. Subjects can report the start of activity (as in~\cite{dong2014detecting,ganti2010multisensor,khan2014activity}). They can manually edit events in a daily calendar that included automatically recognized contexts (similar to~\cite{consolvo2008activity}). 
We treat only the manual corrections or additions as ground truth; the automated predictions act as cues, to help the subjects recall their context (as suggested in~\cite{rahman2014towards}).

The main contribution of this work is the emphasis on \emph{in-the-wild} conditions, as mentioned in ``previous work'':
\begin{enumerate}
	\item Naturally used devices. Subjects used their own personal phones, and a smartwatch that we provided.
	\item Unconstrained device placement. Subjects were free to carry their phone in any way that was convenient to them.
	\item Natural environment. Subjects collected data in their own regular environment for about a week.
	\item Natural behavioral content. No script or tasks were given. We did not target a specific set of activities. Instead, the context labels that we analyze came \emph{from the data}, as the subjects engaged in their routine and applied any relevant labels (from the large menu) that fit what they were doing.
\end{enumerate}

Recognizing context in-the-wild is more challenging, compared to controlled conditions, because of the large variability in real-life.
Diversity in phone devices and sensor hardware has an effect on the measurements~\cite{stisen2015smart}. Our data represents both iPhone and Android, including many varieties of devices.
Variability in behavioral content is clearly visible in the ground truth labels of out data, including combinations like \{Running, Outside, Exercise, Talking, With friends\}, \{Running, Indoors, Exercise, At the gym, Phone on table\}, \{Sitting, Indoors, At home, Watching TV, Eating, Phone on table\}, \{Sitting, At a restaurant, Drinking (alcohol), Talking, Eating\}, \{Sitting, On a bus, Phone in pocket, Talking, With friends\}, \{On a bus, Standing\}. Such variability was missed in works that defined behavior with a small set of mutually exclusive activities. Variability in manner or style (\eg~different gaits) is less visible, but is still captured in our sensor measurements. Such variability can easily be missed in scripted experiments or if restricting how to use devices. Our analysis demonstrates the difficulty in resolving context in-the-wild, and the importance of using complementary sensing modalities. We show that everyday devices, in their natural usage, can capture information about a wide range of behavioral attributes.


\section*{Context recognition system\label{sec:the_system}}

Figure~\ref{fig:principal} illustrates the flow of our recognition system. The system is based on measurements from five sensors in a smartphone: accelerometer (Acc), gyroscope (Gyro), location (Loc), audio (Aud), and phone state (PS), as well as accelerometer measurements from a smartwatch (WAcc). For a given minute, the system samples measurements from these six sensors and the task is to detect the combination of relevant context labels (Figure~\ref{fig:principal}~(A)), \ie~declare for each label $l$ a binary decision: $y_l=1$ (the label is relevant to this minute) or $y_l=0$ (not relevant).

For this paper we opted for simple computational methods, based on linear classifiers and basic heuristics for sensor fusion. We model each label separately and treat every minute as an independent example.
We include time-of-day as part of the PS features, but we do not model the behavioral time-series throughout the day.
The goal of this paper is to show the potential of context recognition \emph{in-the-wild}, and to establish a baseline. Future papers will use non-linear methods, dynamic-context models, and interaction among labels.

\textbf{Single-sensor classifiers} use sensor-specific features and help us understand how informative each sensor can be, independently of the other sensors, for a given context label (Figure~\ref{fig:principal}~(B)). We use logistic regression --- a linear classifier that outputs a continuous value (interpreted as probability) in addition to the binary decision. This is helpful for sensor fusion.
The following procedure was performed for a given sensor $s$ and a given label $l$:
(a) For each example, compute a $d_s$-dimensional feature vector $x_s$. Each sensor has a different set of relevant features. 
(b) Standardize each feature by subtracting mean and dividing by standard deviation (these statistics are estimated on the training set).
(c) Learn a $d_s$-dimensional logistic regression classifier from the training set.
(d) Apply the logistic regression classifier to a test example to obtain a binary classification $y_l$ and probability value $P(y_l=1|x_s)$.
To overcome the imbalance between the positive class and the negative class we applied balanced class weights (inversely proportional to the class frequency in the training set).

At this point, it is possible introduce some domain knowledge and assign appropriate sensors to certain labels. For example, the watch accelerometer can be a good indicator for specific hand-motion activities, like ``washing dishes'', while audio might better predict environmental contexts like ``in class'' or ``at a party''. These design decision are not always obvious, so we continue with sensor-fusion methods that can learn the best predictors from data.

\subsection*{Sensor fusion\label{subsec:sensor_fusion}}
Our system further combines information from $N$ different sensors. We propose three alternative ways.

\begin{figure*}
	\centering
	\includegraphics[width=0.85\textwidth]{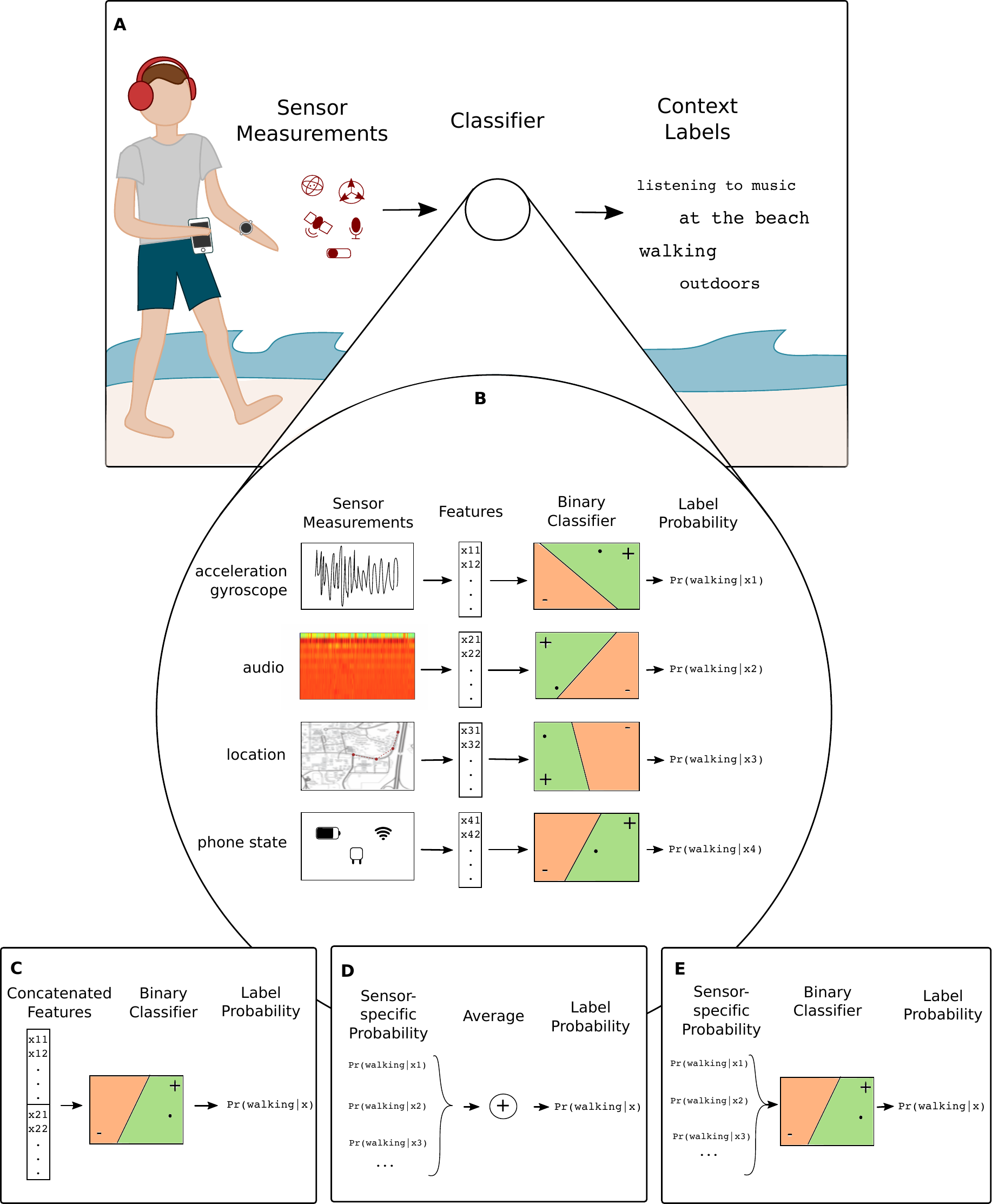}
	\caption{Context recognition system. (A) While the person is engaged in natural behavior the system uses sensor measurements from the smartphone and smartwatch to automatically recognize the person's detailed context.
	(B) Single-sensor classifiers. Appropriate features are extracted from each sensor. For a given context label, classification can be done based on each sensor independently.
	(C) Under Early fusion (EF), features from multiple sensors are concatenated to form a long feature vector. 
	(D) Late fusion using averaging (LFA) simply averages the output probabilities of the single-sensor classifiers. 
	(E) Late fusion with learned weights (LFL) learns how much to ``listen'' to each sensor when making the final classification.}
\label{fig:principal}
\end{figure*}

\textbf{Early fusion (EF)} classifiers combine information from multiple sensors prior to the classification stage (Figure~\ref{fig:principal}~(C)).
The following procedure was performed for a given label $l$:
(a) Start with the sensor-specific feature vectors $\{x_s\}_{s=1}^N$.
(b) Concatenate the (standardized) sensor-specific feature vectors into a single vector $x$ of dimension $d = \sum_{s=1}^{N}{d_s}$
(c) Learn a $d$-dimensional logistic regression classifier from the training set.
(d) Apply the logistic regression classifier to a test example to obtain a binary classification $y_l$ and probability value $P(y_l=1|x)$.

\textbf{Late fusion classifiers.}
We use ensemble methods to combine the predictions of the $N$ single-sensor classifiers. We chose to combine the probability outputs $P(y_l=1|x_s)$, and not the binary decisions, to take into account the ``confidence'' of each of the $N$ classifiers and avoid over-influence of irrelevant sensors. We explore two methods for late fusion:\\
\textbf{Late fusion using average probability (LFA)}  (Figure~\ref{fig:principal}~(D)) applies a simple bagging heuristic and averages the probability values from all the single-sensor classifiers to obtain a final ``probability'' value, \ie, $P(y_l=1|x_1, x_2, \dots, x_{N})=
\frac{1}{N}\sum_{s=1}^{N}{P(y_l=1|x_s)}$. LFA declares ``yes'' if the average probability is larger than half. 
No additional training is performed after the single-sensor classifiers are learned.
This method grants equal weight to each sensor, hoping that informative sensors will classify with higher confidence (probability close to 0 or close to 1) and will influence the final decision more than irrelevant sensors (which will hopefully predict with probability close to 0.5).

As mentioned earlier, some sensors may be consistently better suited for certain labels. As a flexible alternative to deciding apriori how to assign sensors to labels, we can let sensor-weights be learned from data.
\textbf{Late fusion using learned weights (LFL)}  (Figure~\ref{fig:principal}~(E)) is a second type of late fusion that places varying weight on each sensor. This method learns a second layer of $N$-dimensional logistic regression model. The second layer's input is the $N$ probability outputs of the single-sensor models, and the output is a final decision $y_l$.

\section*{Data collection\label{sec:data}}
\begin{figure*}
	\centering
	\includegraphics[width=0.75\textwidth]{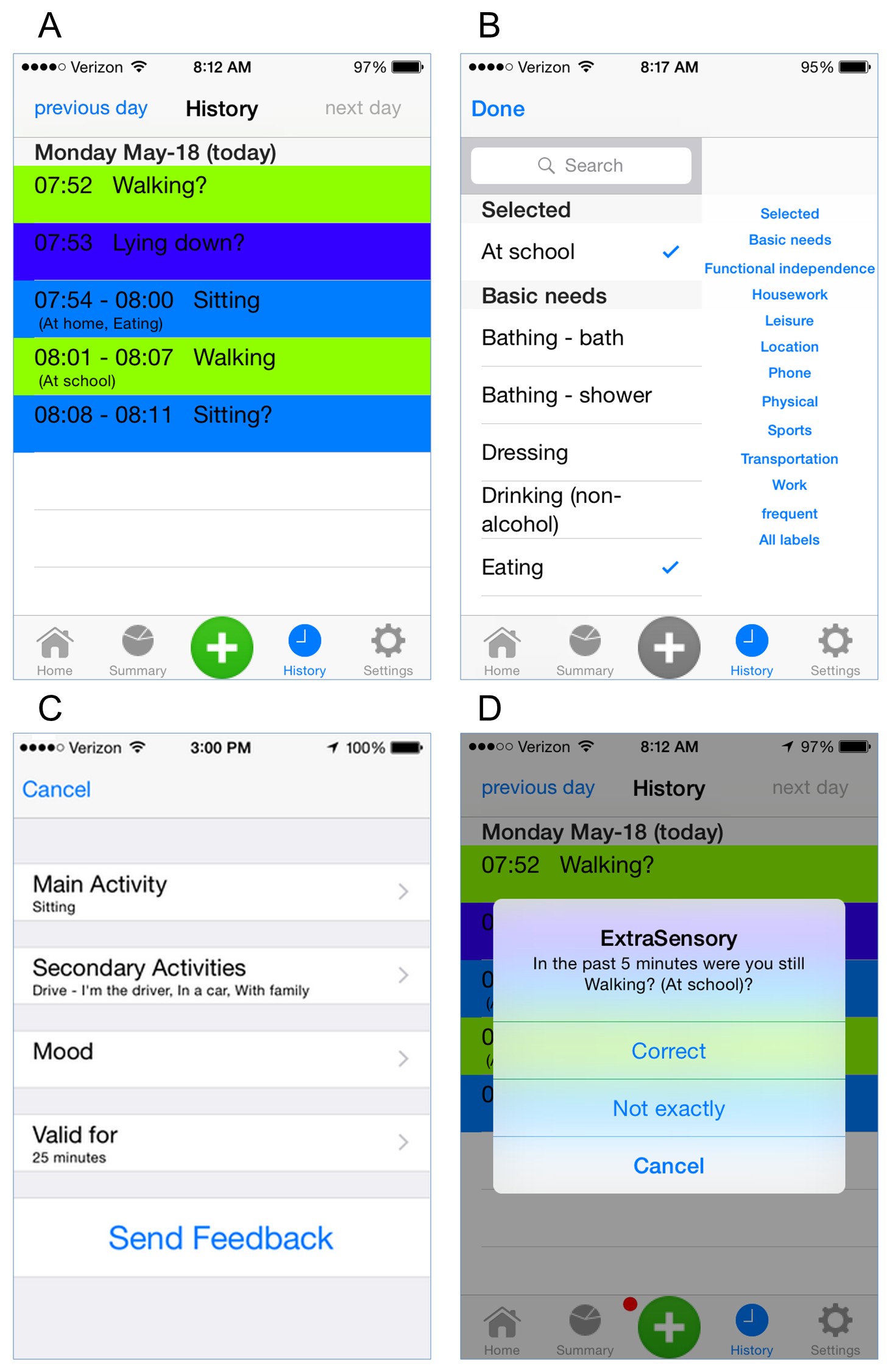}
	\caption{Screenshots from the ExtraSensory mobile application (iPhone version). (A) On the history tab the user can see a daily log of activities and contexts. The server sends real-time body-state predictions (based on preliminary training data from two iPhone users --- the researchers). These predictions appear with question marks and help the user organize the log and recall when their activity may have changed. The user can update the history records' labels, add secondary labels like ``at home'' and ``eating'', merge consecutive records into a longer period, and split records.	(B) The label selection menu is indexed by topics and a ``frequently use'' link to make it easier for the user to select quickly. (C) Using the ``active feedback'' button the user can report they will be engaged in a specific context starting immediately and valid for a set period of time. (D) Periodic notifications remind the user to provide labels. If the user is engaged in the same context as they recently reported they simply need to reply ``correct'' to the question. If any element of the context has changed they can press ``not exactly'' and be directed to a screen where they can update the labels of the recent time period. These notifications appear on the watch as well, which enables easier responses.}
\label{fig:mobileApp}
\end{figure*}

For the purpose of large-scale data collection, we developed a mobile application (app) called \emph{ExtraSensory}, with versions for both iPhone and Android smartphones, and a companion application for the Pebble smartwatch that integrates with both. The app was used to collect both sensor measurements and ground truth context labels. Every minute the app records a 20sec window of sensor measurements from the phone and watch. Within that window, the time samples of different sensors are not guaranteed to be exactly aligned. The flexible user interface provided the user with many mechanisms to self-report the relevant context labels and cover long behavioral time with minimal effort and time of interaction with the app (Figure~\ref{fig:mobileApp}).

Sixty subjects (users) were recruited using fliers posted around the UC San Diego campus and campus-based email lists. 34 of the subjects were iPhone users, with iPhone devices of generations from iPhone4 to iPhone6 and with operating system (iOS) versions 7, 8 and 9. 26 subjects were Android users, with various devices (Samsung, Nexus, Motorola, Sony, HTC, Amazon Fire-Phone, and Plus-One). The subjects were from diverse ethnic backgrounds (self-defined), including Chinese, Mexican, Indian, Caucasian, African-American, and more. The majority of the subjects (93\%) were right handed, and chose to wear the smartwatch on their left wrist. The dataset is homogeneous with regard to occupation; almost all the subjects were students or research assistants. 34 subjects were female and 26 were male. Table~\ref{tab:subjectStats} describes additional subject characteristics.
We installed the app on each subject's personal phone and provided the watch to the subject (56 out of the 60 agreed to wear the watch). The subject then engaged in their usual behaviors for approximately a week, while keeping the app running in the background on their phone as much as possible and convenient. The subject was asked to report as many labels as possible without interfering too much with their natural behavior. They were free to remove the watch whenever they wanted and were asked to turn off the watch-app when they were not wearing it. Basic compensation of $\$40$ was given to each subject, with additional incentive of up to $\$35$ that depended on the amount of labeled data they provided.
 
The resulted \emph{ExtraSensory Dataset}, contains a total of 308,320 labeled examples (minutes) from sixty users. Table~\ref{tab:subjectStats} details statistics (over 60 subjects) about the amount of data collected. Not all the sensors were available at all times. Table~\ref{tab:sensorStats} specifies details on the sensors. The dataset is publicly available and researchers are encouraged to use it for developing and comparing context recognition methods (http://extrasensory.ucsd.edu).

\begin{table}[h]
	\centering
	\begin{tabular}{r|c|c}
		& Range & Mean (SD)\\
		\hline
		Age (years) & 18--42 & 24.7 (5.6)\\
		Height (cm) & 145--188 & 171 (9)\\
		Weight (kg) & 50--93 & 66 (11)\\
		Body mass index (kg/m$^2$) & 18--32 & 23 (3)\\
		\hline
		Labeled examples & 685--9706 & 5139 (2332) \\
		Additional unlabeled examples & 2--6218 & 1150 (1246) \\
		Average applied labels per example & 1.1--9.7 & 3.8 (1.4) \\
		Participation duration (days) & 2.9--28.1 & 7.6 (3.2) 
	\end{tabular}
	\caption{Statistics over the 60 users in the dataset (SD:\ standard deviation).}
\label{tab:subjectStats}
\end{table}

\begin{table}[h]
	\centering
	\begin{tabular}{l|l|c|c}
		Sensor & raw measurements & examples & users  \\
		\hline
		Accelerometer & 3-axis (40Hz) & 308,306 & 60 \\
		Gyroscope & 3-axis (40Hz) & 291,883 & 57 \\
		Magnetometer & 3-axis (40Hz) & 282,527 & 58 \\
		Watch Accelerometer & 3-axis (25Hz) & 210,716 & 56 \\
		Watch Compass & heading angle (var) & 126,781 & 53 \\
		Location & long-lat-alt (var) & 273,737 & 58 \\
		Location precomputed & location variability (1pe) & 263,899 & 58 \\
		Audio & 13MFCC (46ms frames) & 302,177 & 60 \\
		Audio power & 1pe & 303,877 & 60 \\
		Phone State & 1pe & 308,320 & 60 \\
		Low frequency sensors & 1pe & 308,312 & 60 \\
		\hline
		Core & & 176,941 & 51 \\
	\end{tabular}
	\caption{The sensors in the dataset. For each sensor, details of the raw measurements, the number of examples with measurements from that sensor and the number of users with measurements from that sensor. ``Core'' represents examples that have measurements from all six core sensors that are analyzed in this paper (Acc, Gyro, WAcc, Loc, Aud and PS). ``1pe'' means sampled once per example. ``var'' means variable sampling rate --- gathering updates whenever the value changes.}
\label{tab:sensorStats}
\end{table}

\section*{Evaluation and results\label{sec:results}}
We evaluated classification performance using five-fold cross validation: each fold has 48 users in the training set and the other twelve users in the test set. We also conducted leave-one-user-out experiments (LOO). To measure performance, classification accuracy is a misleading metric because of imbalanced data; for a rare label that appears in 1$\%$ of the test set, a trivial classifier that always declares ``no'' will achieve 99$\%$ accuracy but is completely useless. It is important to consider competing metrics, like sensitivity and specificity. A common approach is to observe sensitivity (recall) against precision, or to calculate their harmonic mean (F1). However, precision and F1 are less fitting, since they are very sensitive to how rare labels are. Chance level can be arbitrarily small, and when averaging precision or F1 over many labels, certain labels will unfairly dominate the score. Additionally, the self-reported data may be noisy, possibly including cases where a label was actually relevant, but was not reported by the subject. Precision and F1 will be too sensitive to such cases. Unlike F1, the balanced accuracy, BA=0.5*(sensitivity+specificity), does not suffer from these issues, and can serve as a convenient objective that fairly balances competing metrics.

\begin{figure*}
	\centering
	\includegraphics[width=0.9\textwidth]{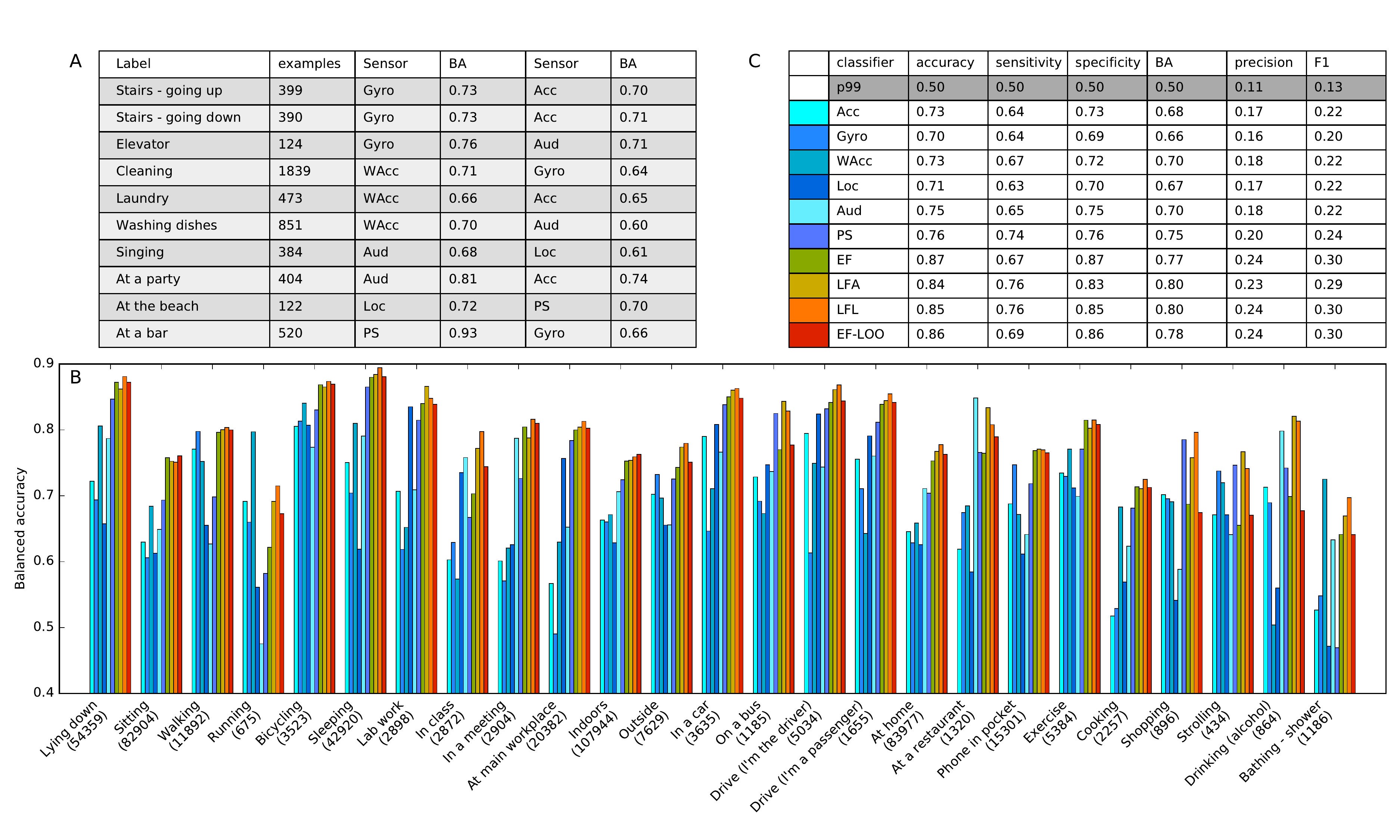}
	\caption{Overall performance of the single-sensor classifiers (Acc, Gyro, WAcc, Loc, Aud and PS) and the sensor-fusion classifiers (EF, LFA, LFL and EF-LOO).
	(A) Specific labels that had few examples with first and second guess of sensors that intuitively seem relevant and the BA score of the corresponding single-sensor classifiers.
	(B) BA scores for selected labels from diverse domains (number of examples in parenthesis). Color legend is in table (C).
	(C) Average performance metrics over the 25 context labels from B. All average scores were well above the p99 value, which marks the 99th percentile of random score --- scores above the p99 value have less than 1\% probability of being achieved randomly (p99 was estimated from 100 random simulations).}
\label{fig:barsAndAverage}
\end{figure*}

First, we assess the potential of single sensors. Figure~\ref{fig:barsAndAverage}~(A) shows some specific context labels for which relatively few examples were collected. If we pick our first (sometimes second) guess of relevant sensor we can achieve reasonable recognition of these contexts.

Next, to see if we can do better, we evaluate the three sensor-fusion methods described in ``Sensor fusion'' and compare them to single-sensor classifiers. Figure~\ref{fig:barsAndAverage}~(B) shows performance for 25 labels from diverse context domains. In most cases sensor-fusion managed to match the best fitting single-sensor. The system learned from data how to best utilize the different sensors, without the need of a human to guide it, which can be useful for scalable systems, where the researcher does not necessarily know which sensor to trust for which label. Furthermore, in many cases sensor-fusion improved performance, compared to the best single-sensor, meaning that there is complementary information in different sensors. We see the overall advantage of multi-sensor systems over single-sensor systems, shown by the average performance of the different systems in Figure~\ref{fig:barsAndAverage}~(C). The three sensor-fusion alternatives seem to perform similarly well, with LFL slightly ahead. The selection of a sensor-fusion method can be guided by the training data available to the researcher. When having plenty of labeled examples that have all six sensors available, the simple EF system can work. Otherwise, late fusion will be more fitting, still having plenty of data to train each single-sensor classifier alone.
Leave-one-user-out results are consistent with 5-fold evaluation (figure~\ref{fig:barsAndAverage} shows LOO results for the EF system, marked ``EF-LOO''). For some labels, like ``running'', the system benefited from the larger training set in the LOO evaluation. Full per-label results are provided in supplemental material.

\subsection*{Why does sensor fusion help?}
\begin{figure*}
	\centering
	\includegraphics[width=\textwidth]{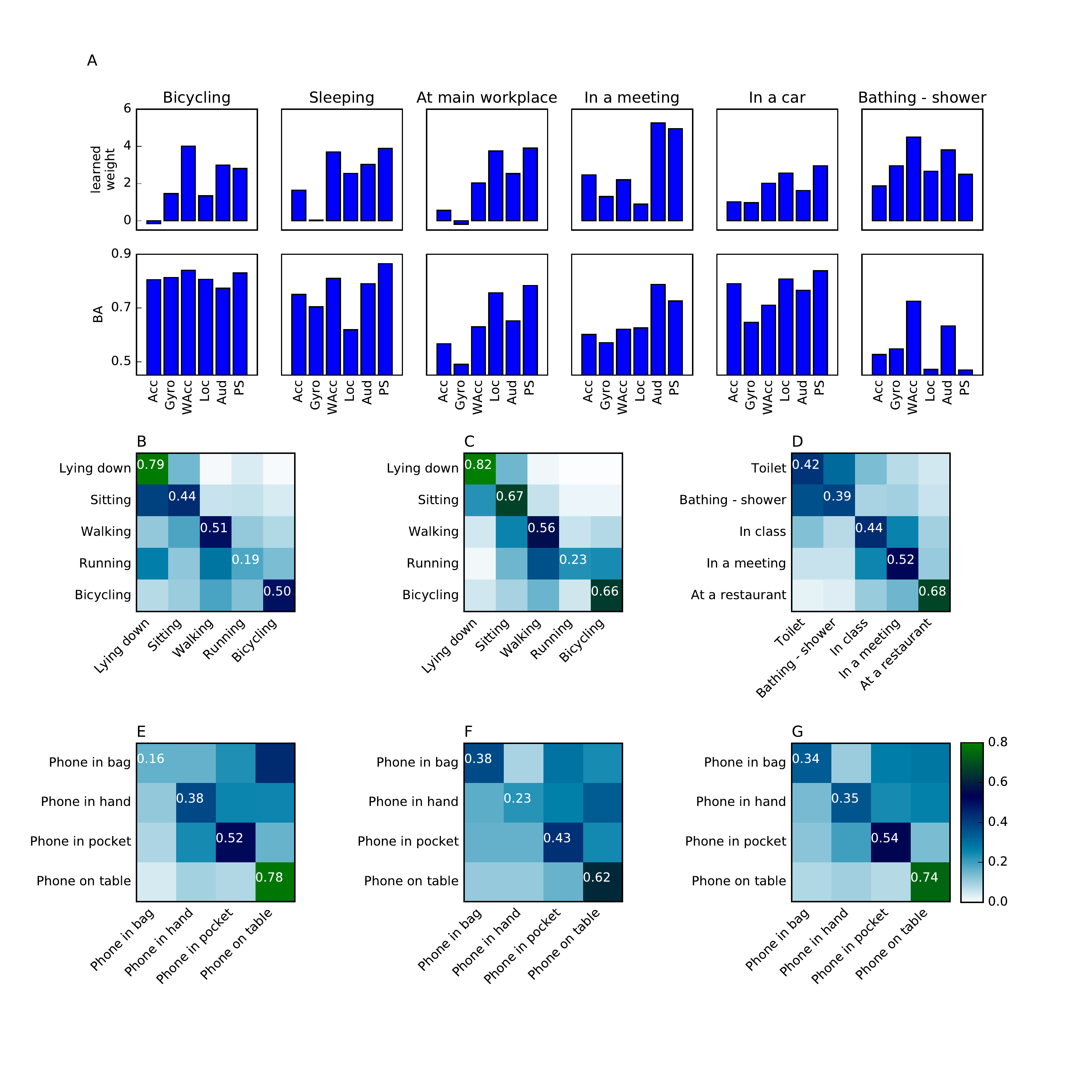}
	\caption{Why sensor fusion helps recognition. (A) The bottom row shows the overall performance (BA) of each single-sensor classifier and the top row shows the weights that the LFL classifier learned to assign to each sensor (taken from the first cross validation fold). (B--G) Confusion matrices for subsets of mutually exclusive labels. Rows represent ground truth labels and columns represent predicted labels. Rows are normalized so that a cell in row $i$ and column $j$ displays the proportion of examples of class $i$ that were assigned to class $j$. The correct classification rates (main diagonal) are also marked numerically. The sensors used are: (B) Acc and Gyro; (C) Acc, Gyro and WAcc; (D) Aud; (E) Acc, Gyro and WAcc; (F) Aud; (G) Acc, Gyro, WAcc and Aud.}
\label{fig:weightsAndConfmats}
\end{figure*}

The performance of single-sensor classifiers on selected labels (Figure~\ref{fig:weightsAndConfmats}~(A)) demonstrates the advantage of having sensors of different modalities. As expected, for detecting sleeping, the watch is more informative than the phone's motion sensors (Acc, Gyro) --- during sleep the phone may be lying motionless on a nightstand, while the watch records wrist movements. Similarly, contexts such as ``shower'' or ``in a meeting'' have unique acoustic signatures (running water, voices) that allow the audio-based classifier to perform well.
When showering, it is reasonable that the phone will be in a different room than the person, in which case the watch is an important indicator of the activity.
Figure~\ref{fig:weightsAndConfmats}~(A) demonstrates that the LFL method assigns reasonable weights to the six sensors --- sensors that perform more strongly for a given label are given higher weight.

Investigating where misclassification occurs helps to understand the predictive ability of the system.
Figure~\ref{fig:weightsAndConfmats}~(B--G) shows confusion matrices that depict misclassification rates between related context labels. 
For example, a classifier using the phone's motion sensors (Acc and Gyro) (Figure~\ref{fig:weightsAndConfmats}~(B)) to discriminate between body movement/posture states confuses even dissimilar labels (``running'' \vs~``lying down''). Such errors arise in natural, unconstrained behavior; in-the-wild, people do not always carry their phone in their pocket --- subjects were sometimes running on a treadmill with their phone next to them, motionless. The watch can help in such situations --- when the watch accelerometer features were added to the classifier (Figure~\ref{fig:weightsAndConfmats}~(C)), the confusion between activities was reduced.

The audio signal from the smartphone (Figure~\ref{fig:weightsAndConfmats}~(D)) is informative for labels related to the environmental context. We see a hierarchy of misclassification: while there is some confusion between labels that share similar acoustic properties (``toilet''~\vs~``shower'', ``class''~\vs~``meeting''), there is a sharper distinction between label groups from different domains (``toilet or shower'' \vs~``class or meeting'' \vs~``restaurant'').

The phone placement itself provides cues about the user's activity; when the phone is lying on a table it is more likely the user is showering than walking to work. The ability to recognize the phone's position will improve overall context recognition. A single modality is not sufficient to fully identify phone position. A classifier based on motion sensors is sensitive to movement, so when the phone was in a bag (possibly motionless) it was often mistaken for being on a table (Figure~\ref{fig:weightsAndConfmats}~(E)). On the other hand, a classifier using audio alone is more sensitive to whether the phone is enclosed or exposed to environmental sounds, so with this classifier cases of ``phone in bag'' were mistaken for ``phone in pocket'' and ``phone in hand'' was often mislabeled as ``phone on table'' (Figure~\ref{fig:weightsAndConfmats}~(F)). However, by combining motion and audio modalities, the classifier synthesized these two dimensions of discrimination to better recognize phone position (Figure~\ref{fig:weightsAndConfmats}~(G)). These examples demonstrate the large variability in behavior in-the-wild and highlight the utility of fusing multi-modal sensors.

\section*{User personalization}
People move, behave and uses their phone in different manners. A system that is fine-tuned to its specific user may outperform a more general model. To explore the potential of personalization we performed experiments with a single test user. We compared three models: (1) universal (trained on data from other users), (2) individual (trained on half of the data from the same test user) and (3) adapted (merges both). We tested the three models on the same unseen data.

\begin{figure*}
	\centering
	\includegraphics[width=\textwidth]{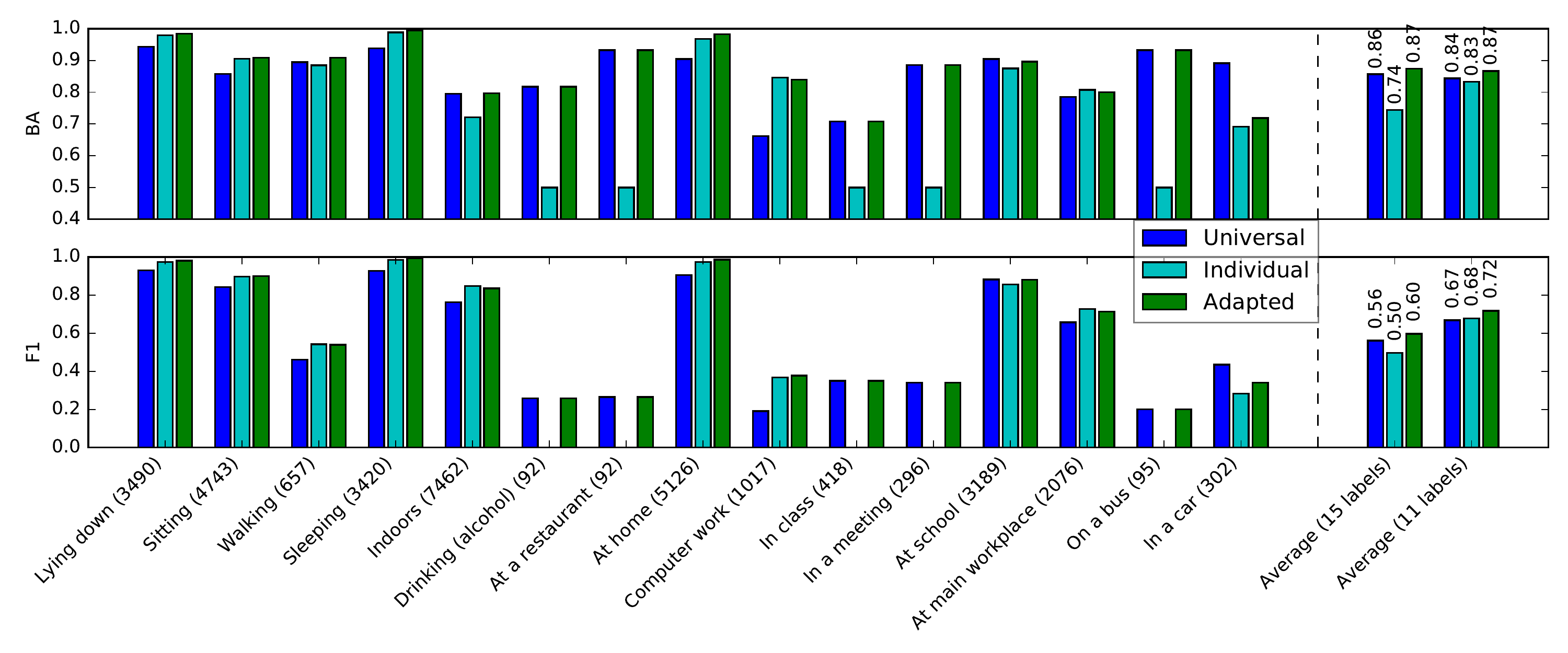}
	\caption{User adaptation performance. Balanced accuracy and F1 score were assessed for a single user. The x-axis denotes the labels with the total number of examples for the user. The ``universal'' model was trained on data from other users, the ``individual'' model was trained on data from the same user and the ``adapted'' model combines the universal and individual models using LFA.\@ The bars on the right hand side of each plot present the scores averaged over the 15 tested labels, and averaged over 11 labels that had over 300 examples.}
\label{fig:personalization}
\end{figure*}

Figure~\ref{fig:personalization} shows the results of these experiments.
The universal model demonstrates good performance. This shows the basic ability of a trained system to work well for a new unseen user.
As suspected, the individual model performed better than the universal model for labels that had many individual examples (``lying down'', ``sitting'', ``sleeping'', ``at home'', ``computer work'', and ``at main workplace''). However, the individual user is missing data for many context labels. For other labels there are only a limited number of examples a new user can acquire in a few ``training'' days, which risks over-fitting to these few examples. In such cases a universal model is better, having been trained on plenty of data from many users.
The optimal solution is to benefit from both universal and individual data: the user-adapted model shows overall improvement in recognition performance, even among the labels that had over 300 examples for the test user. LFA is a simple heuristic that manages to demonstrate this advantage. For each label, when there is not enough data to train an individual model, the adapted model relies only on the universal model. When there is enough data to train an individual model, the adapted model ``listens'' to the universal and individual models, in some cases achieving better performance than either model on its own (\eg~``sleeping'', ``at home'').

In practical systems, the logistics of implementing personalization may not be an obvious task. For medical applications, the clinician or patient may decide that their cause is important and worth dedicating some effort to provide individual labeled data for a few days, in order to better adapt the model. However, in commercial applications the users (clients) may not be motivated to invest the extra effort in labeling. In such cases semi-supervised methods can still be used to make the most of unlabeled data from the individual user and personalize the model.

\section*{Conclusions}
Our novel data collection brings about behavioral variability in-the-wild that is underrepresented in controlled studies. This makes context recognition a harder challenge compared to previous scenarios, hence accuracy levels in-the-wild are lower than those reported in experiments that had some restrictions on behavioral conditions.
We demonstrate that everyday devices, in their normal unconstrained usage, carry information about the person's natural behavioral context. We describe a baseline system, suggest three simple methods for sensor fusion, and reinforce previous findings that showed the advantage of fusing multi-modal sensors. We demonstrate how the sensing modalities complement each other, and help resolve contexts that arise with uncontrolled behavior (\eg~running on treadmill with phone on table, motionless).

Combinatorial representation of behavior is very flexible. A well trained system has the potential to correctly recognize a new specific situation (combination of labels) that did not appear in the training. 
To broaden the range of contexts, researchers can either use supervised methods and focus on newly added target labels when collecting extra data, or use unsupervised methods to discover complex behaviors as common combinations or sequences of simpler contexts~\cite{seiter2014discovery}.
The labels in our work were interpreted in a subjective manner. The same location may be considered as ``school'' for one subject and ``workplace'' for another. We did not tell the subjects how we define ``walking'' or ``eating'' in order to capture the full scope of what people consider eating. Domain-expert researchers may decide to define labels clearly to subjects or use more specific labels like ``eating a meal'' and ``snacking''.

\section*{Future directions}


New technologies and original solutions for collecting labels in-the-wild are required to reduce annotation load from study subjects and increase reliability of labeling. Online learning can be used to keep improving real-time recognition, which will require less label-correcting effort from new research subjects. Active learning can be utilized to collect data in scale, while sparsely probing subjects to provide annotations.
In parallel, semi-supervised methods can be used to make the best out of plenty unlabeled data (which is easy to collect) and reduce the dependence on labeled examples to a minimum.\\
The public dataset we collected provides a platform to develop and evaluate these methods, as well as explore feature extraction, inter-label interaction, time series modeling and other directions that will improve context recognition.

\ifCLASSOPTIONcaptionsoff
  \newpage
\fi

\bibliographystyle{IEEEtran}
\bibliography{extrasensory_references}

\begin{IEEEbiography}[{\includegraphics[width=1in,height=1.25in,clip,keepaspectratio]{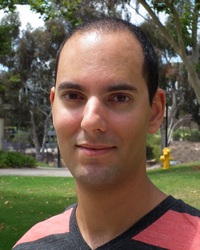}}]{Yonatan Vaizman}
Yonatan Vaizman is a Ph.D. candidate in the department of Electrical and Computer Engineering at UC San Diego. His fields of interest are machine learning, signal processing, and artificial intelligence. In his research, he applies methods from these fields to music recommendation, mobile sensor processing, and human behavior recognition. He received his B.S. degree in Computer Science and Computational Biology from the Hebrew University in Jerusalem, Israel, and his M.S. degree in Electrical Engineering from UC San Diego.
\end{IEEEbiography}

\begin{IEEEbiography}[{\includegraphics[width=1in,height=1.25in,clip,keepaspectratio]{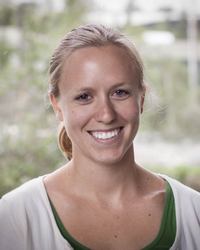}}]{Katherine Ellis}
Katherine Ellis is a Research Scientist at Amazon. Her research interests are in applications of machine learning to physical activity measurement, mobile health, social network analysis and music recommendation. 
She received a B.S. degree in electrical engineering from the University of Southern California and M.S. and Ph.D. degrees in electrical engineering from UC San Diego. 
She is an IEEE member and an ACM member.
\end{IEEEbiography}


\begin{IEEEbiography}[{\includegraphics[width=1in,height=1.25in,clip,keepaspectratio]{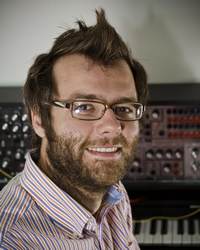}}]{Gert Lanckriet}
Gert Lanckriet is a Principal Applied Scientist at Amazon, and Professor of Electrical and Computer Engineering at UC San Diego. His interests are in data science, on the interplay between machine learning, applied statistics, and large-scale optimization, with applications to music and video search and recommendation, multimedia, and personalized, mobile health.

He was awarded the SIAM Optimization Prize in 2008 and is the recipient of a Hellman Fellowship, an IBM Faculty Award, an NSF CAREER Award, and an Alfred P. Sloan Foundation Research Fellowship. In 2011, MIT Technology Review named him one of the 35 top young technology innovators in the world (TR35). In 2014, he received the Best Ten-Year Paper Award at the International Conference on Machine Learning.

He received a master’s degree in electrical engineering from the Katholieke Universiteit Leuven, Belgium, and M.S. and Ph.D. degrees in electrical engineering and computer science from UC Berkeley. He is a senior IEEE member and an ACM member.
\end{IEEEbiography}

\clearpage

\setcounter{figure}{0}
\makeatletter 
\renewcommand{\thefigure}{S\arabic{figure}}

\setcounter{table}{0}
\makeatletter 
\renewcommand{\thetable}{S\arabic{table}}

\section*{Supplementary material}

Supplementary material has technical details about the following components of the work:
\begin{itemize}
	\item Mobile app
	\item Data collection procedure
	\item Sensor measurements
	\item Extracted features
	\item Label processing
	\item Classification methods
	\item Performance evaluation
	\item User personalization assessment
	\item Detailed results tables
\end{itemize}

\subsection*{Mobile app}
For the purpose of data collection in a large scale we developed a mobile application called \emph{ExtraSensory}, with versions for both iPhone and Android smartphones, and a companion application for the Pebble smartwatch that integrates with both. The app was used for supervised data collection, meaning it collects both sensor measurements and ground truth context labels.
The app is scheduled every minute to automatically record measurements for 20 seconds from the sensors. Sensors are sampled in frequencies appropriate for their domain, and include motion-responsive sensors, location services, audio, environment sensors, as well as bits of information about the phone's state. When the watch is available (within Bluetooth range and paired with the phone) measurements from the watch are also collected by the app during the 20 second recording session. More details about the sensors are provided in ``Sensor measurements''. At the end of the 20 second recording session the measurements are bundled in a zip file and sent through the internet (if a WiFi network is available) to our lab's server, which runs a quick calculation and replies with an initial prediction of the activity (\eg~sitting, walking, running). All communication between the app and the server is secure and encrypted, and identified only by a unique universal identifier (UUID) that was randomly generated for each user.

In addition to collecting sensor measurements, the app's interface provides several mechanisms for the user to report labels about their context. This was a crucial part of the research design and we had to overcome a basic trade-off: on one hand we wanted to collect ground truth labels for as many examples (minutes) as possible and with much detail (combination of all the relevant context labels). On the other hand we did not want the subject to interact with the app every minute to report labels, both because it would be an extreme inconvenience for the subject and because it would impact the natural behavior of the subject and miss the point of collecting data in-the-wild. To balance this trade-off, we designed a flexible interface that helps minimize the user-app interaction time. The following label-reporting mechanisms were included:
\begin{itemize}
	\item A history journal presents the user's activities chronologically as a calendar and enables the user to easily edit the context labels of time ranges in the past (up to one day in the past). The user can easily merge a sequence of minutes to a single ``event'' with the same context labels, or split a calendar event to describe a change in context. See Figure~\ref{fig:mobileApp}~(A). The real-time predictions from the server assist the user to recall when their activity  changed --- consecutive minutes with the same prediction from the server are merged to a single item on the history calendar.
	\item The user can also initiate active feedback by reporting labels describing their context in the immediate future (starting ``now'' for up to half an hour in the future).  See Figure~\ref{fig:mobileApp}~(B).
	\item Every $x$ minutes (by default, $x$ is 10 minutes, but can be set by the user) the app presents a notification to remind the user to provide labels. If the the user has recently provided labels, the notification asks whether the user was still engaged in the same activities --- allowing for a  quick and easy response if the answer is ``yes''. See Figure~\ref{fig:mobileApp}~(C).
	\item The notifications also appear on the smartwatch, allowing for an easier response with a click of a button on the watch, without using the phone itself.
	\item When selecting labels from the menu, a side-bar index allows quick search of the relevant labels, either by categories (\eg~sports, work, company) or through a ``frequently used labels'' menu, which presents labels that the user has applied in the past. The category in which a label was presented in the menu does not matter, and a label can appear under different categories (\eg~``skateboarding'' appears under ``sports'', ``leisure'' and ``transportation'') --- the only reason for these categories is to make it easy for the user to find the relevant label quickly.
\end{itemize}

\subsection*{Data collection procedure}
The study's research plan and consent form were approved by the university's institutional review board (IRB). Human subjects were recruited for the study via fliers across campus, university mailing lists and word of mouth. Every subject read and signed the consent form. The researchers installed the app on each subject's personal phone (to maximize authenticity of natural behavior). The subject then engaged in their usual behaviors for approximately a week, while keeping the app running in the background on their phone as much as was possible and convenient. The subject was asked to report as many labels as possible without interfering too much with their natural behavior. 
Subjects varied in their level of rigorousness with respect to providing labels: some wanted to be very precise (with specific detailed combinations of labels, and trying to keep minute-to-minute precision) and others tended to be less specific and to dedicate less effort. The subjects who used the watch, which we supplied them with, were told that it is fine to get it wet (wash hands, shower) but not submerge it (swimming). They were also asked to turn off the watch app whenever they removed the watch from their wrist and to turn it back on when they wore the watch --- so we can generally assume that whenever watch measurements are available they were taken from the subject's wrist.

Using the app consumes the phone's battery more quickly than normal. To make up for this, the researchers provided participants with an external portable battery, which provides one extra charge during the day.
The researchers also provided the subject with the Pebble smartwatch (56 of the subjects agreed to use the watch). The external battery and the smartwatch were returned at the end of the study.
Each subject was compensated for their participation. The basic compensation was in the amount of $\$40$, and an additional amount was calculated based on the amount of labeled data that the subject contributed (as an incentive to encourage reporting many labels). The total compensation per subject was between $\$40$ and $\$75$.

\subsubsection*{Technical difficulties}
During the development of the iPhone app, there were releases of new iOS versions that caused the app to not work well and required us to change the code.\\
Since subjects used their personal phones, the app had to handle different devices, and in the Android case, different makers.
For some of the Android users when we installed the app we noticed it didn't work well. In three cases the workaround was to install a slightly different version of the app that didn't use the gyroscope. After installing the changed app and making sure it works those users began collecting records (without gyroscope measurements).\\
On top of the dataset's 60 subjects, there were four more subjects that participated and received the basic compensation, but whose data was not included in the dataset. For two of them the app didn't work well on their devices. The other two were too stressed or otherwise occupied during participation, and produced too little and un-reliable labels, so we decided to discard their data.

\subsection*{Sensor measurements}
Raw sensor measurements are provided in the publicly available dataset.\\
\emph{High frequency measurements:}\\
Each sensor (and pseudo-sensor) in the following list was sampled at $40$Hz during the $\sim$20 second recording session to produce a time series of $\sim$800 time points. The sampling relies on the design of the phone's hardware and operating system and the sampling rate was not guaranteed to be accurate (especially for the Android devices). For that reason the time reference of each sample in a time series was also recorded; the differences between consecutive time references were approximately 25 milliseconds.
\begin{itemize}
	\item Accelerometer. Time series of 3-axis vectors of acceleration according to standard axes of phone devices. 
	\item Gyroscope. Time series of 3-axis vectors of rotation rate around each of the phone's 3 axes.
	\item Magnetometer. Time series of 3-axis vectors of the magnetic field. 
	\item Processed signals. Both iPhone and Android operating systems provide processed versions of the signals: The raw acceleration is split to the gravity acceleration (estimated direction of gravity at every moment, the magnitude is always 1G) and the user-generated acceleration (subtraction of the gravity signal from the raw acceleration). For the gyroscope the OS has a calibrated version that attempts to remove drift effects. For the magnetometer the OS has an unbiased version that subtracts the estimated bias of the magnetic field created by the device itself.
\end{itemize}
In this paper we used the raw acceleration signal (which includes the effects of gravity) and the calibrated version of the gyroscope signal.
Acceleration is reported in units of G (gravitational acceleration on the surface of the Earth) on iPhone and in units of m/s\textsuperscript{2} on Android. Before extracting features we converted the Android acceleration measurements to units of G.

\emph{Watch measurements:} From the Pebble smartwatch we collected signals from the two available sensors---accelerometer and compass.
Acceleration was sampled at 25Hz and describes a 3-axis vector of acceleration (in units of mG) relative to the watch's axes-system. The compass does not have a constant sampling rate; it was requested to provide an update of the heading whenever a change of more than one degree was detected. The compass takes some time to calibrate before providing measurements, so some examples that have watch acceleration measurements are missing compass measurements.

\emph{Location measurements:} Both iPhone and Android provide location services (based on a combination of GPS, WiFi and cellular communications). The app samples location data in a non-constant rate, as the location service updates each time a movement is detected. This creates a time series of varying length (sometimes just a single time point in a recording session, sometimes more than 20 points) of location updates. Each update has a relative time reference and the estimated location measurements: latitude coordinate, longitude coordinate, altitude, speed, vertical accuracy and horizontal accuracy (these accuracies describe the range of reasonable error in location). Some of these values may be missing at times (\eg~when the phone is in a place with weak signals).
In addition to the time series of location updates, the app calculates on the phone some basic heuristic location features: standard deviation of latitude values, standard deviation of longitude values, total change of latitude (last value minus first value), total change of longitude, average absolute latitude derivative and average absolute longitude derivative (as proxy to the speed of the user).

To protect our study subjects' privacy (collected examples with label ``at home'' that also include the exact location coordinates may reveal the subject's identity) the app has an option to select a location (typically their home) they would like to disguise. For the subjects that opted to use this option, whenever they were within 500 meters of their chosen location, the app would not send the latitude and longitude coordinates from the current recording (but it would send the other estimated location values such as altitude, speed, as well as the basic heuristic location features).

\emph{Low frequency measurements:} These measurements were sampled just once in a recording session (approximately once per minute). Some of them describe the phone state (PS): app state (foreground/background), WiFi connectivity status, battery status (charging, discharging), battery level, or phone call status.
Other low frequency measurements are taken from sensors built in to the phone, if available: proximity sensor, ambient light, temperature, humidity, air pressure.

\emph{Audio data:} Audio was recorded from the phone's microphone in 22,050 Hz for the duration of the recording session ($\sim$20 seconds). Audio was not available for recording when the phone was being used for a phone call. In order to maintain the privacy of the subjects, the raw audio recording was not sent to the server. Instead, standard audio processing features (Mel Frequency Cepstral Coefficients (MFCC)) were calculated on the phone itself and only the features were sent to the server. The MFCCs were calculated for half-overlapping windows of 2048 samples, based on 40 Mel scaled frequency bands and 13 cepstral coefficients (including the 0\textsuperscript{th} coefficient). As part of the preprocessing of the recorded audio the raw audio signal was normalized to have maximal magnitude of 1 (dividing by the maximum absolute value of the sound wave). This normalizing factor is also sent as a measurement separately from the calculated MFCC features.

\subsection*{Extracted features}
For the experiments in this work we focused on six sensors: accelerometer, gyroscope, watch accelerometer, location, audio and phone state. Other sensors' measurements are available in the public dataset.
Every sensor measures different physical or virtual properties and has a different form of raw measurements. Therefore we designed specific features for each sensor. The published dataset includes files with these features pre-computed for all the users.

\emph{Accelerometer and Gyroscope} (26 features each): Since in natural behavior the phone's position is not controlled we cannot assume it is oriented in a particular way, and it also may be  changing its axes-system with respect to the ground (and with respect to the person). For that reason we had little reason to assume that any of the phone's axes will have a particular coherent correspondence to many behavioral patterns, and we extracted most of the features based on the overall magnitude of the signal.
We calculated the vector magnitude signal as the euclidean norm of the 3-axis acceleration measurement at each point in time, \ie, $a[t] = \sqrt{{a_x[t]}^2+{a_y[t]}^2+{a_z[t]}^2}$.
We extracted the following features:
\begin{itemize}
	\item Nine statistics of the magnitude signal: mean, standard deviation, third moment, fourth moment, 25$^{th}$ percentile, 50$^{th}$ percentile, 75$^{th}$ percentile, value-entropy (entropy calculated from a histogram of quantization of the magnitude values to 20 bins) and time-entropy (entropy calculated from normalizing the magnitude signal and treating it as a probability distribution, which is designed to detect peakiness in time---sudden bursts of magnitude).
	\item Six spectral features of the magnitude signal: log energies in 5 sub-bands (0--0.5Hz, 0.5--1Hz, 1--3Hz, 3--5Hz, $>$5Hz) and spectral entropy.
	\item Two autocorrelation features from the magnitude signal. The average of the magnitude signal (DC component) was subtracted and the autocorrelation function was computed and normalized such that the autocorrelation value at lag 0 will be 1. The highest value after the main lobe was located. The corresponding period (in seconds) was calculated as the dominant periodicity and its normalized autocorrelation value was also extracted.
	\item Nine statistics of the 3-axis time series: the mean and standard deviation of each axis and the 3 inter-axis correlation coefficients.
\end{itemize}

\emph{Watch accelerometer} (46 features): From the watch acceleration we extracted the same 26 features as from the phone accelerometer or gyroscope. Since the watch is positioned in a more controlled way than the phone (it is firmly fixed to the wrist), its axes have a strong meaning (\eg~motion along the x-axis of the watch describes a different kind of movement than motion along the z-axis of the watch). Hence we added 15 more axis-specific features---log energies in the same 5 sub-bands as before, this time calculated for each axis' signal separately. In addition, to account for the changes in watch orientation during the recording we calculated 5 relative-direction features in the following way: we first calculate the cosine-similarity between the acceleration directions of any two time points in the time series (value of 1 meaning same direction, value of -1 meaning opposite directions and value of 0 meaning orthogonal directions). Then we averaged these cosine similarity values in 5 different ranges of time-lag between the compared time points (0--0.5sec, 0.5--1sec, 1--5sec, 5--10sec, $>$10sec).

\emph{Location} (17 features): In this work we used location features that were based only on relative locations, and not on absolute latitude/longitude coordinates. This was in order to avoid over-fitting to our location-homogeneous training set that will not generalize well to the outside world (\eg,~mistakenly learning that a specific location in the UCSD campus always corresponds to ``at work''). Six features were calculated on the phone --- this was in order to have some basic location features in cases where the subjects opted to hide their absolute location. These quick features included standard deviation of latitude, standard deviation of longitude, change in latitude (last value minus first value), change in longitude, average absolute value of derivative of latitude and average absolute value of derivative of longitude.

The transmitted location measurements were further processed to extract the following 11 features: number of updates (indicating how much the location changed during the 20 second recording), log of latitude-range (if latitudes were transmitted), log of longitude-range (if longitudes were transmitted), minimum altitude, maximum altitude, minimum speed, maximum speed, best (lowest) vertical accuracy, best (lowest) horizontal accuracy and 
diameter (maximum distance between two locations in the recording session, in meters).

\emph{Audio} (26 features): From the time series of 13-dimensional MFCC vectors (typically around 400 time frames) we calculated the average and standard deviation of each of the 13 coefficients.

\emph{Phone State} (34 features): For this work we used only the discrete phone state measurements. We represented them with a 26-dimensional one-hot representation---for each property we created a binary indicator for each of the possible values the property can take, plus one indicator denoting missing data. This representation is a redundant coding of the phone state, but it facilitates the use of simple, linear classifiers over this long binary vector representation. The keys were: app state (3 options: active, inactive, background), battery plugged (3 options: AC, USB, wireless), battery state (6 options: unknown, unplugged, not charging, discharging, charging, full), in a phone call (2 options: false, true), ringer mode (3 options: normal, silent no vibrate, silent with vibrate) and WiFi status (3 options: not reachable, reachable via WiFi, reachable via WWAN).

In addition, we added a set of features indicating time-of-day information. We used the timestamp of every example and (using San Diego local time) extracted the hour component (one out of 24 discrete values). In order to get a flexible, useful representation we defined 8 half-overlapping time ranges: midnight-6am, 3am-9am, 6am-midday, 9am-3pm, midday-6pm, 3pm-9pm, 6pm-midnight and 9pm-3am. Then we represented each example's hour with an 8-bit binary value, where 2 bins will be active for 1 relevant time range.

\subsection*{Label processing}
Since the labels are obtained by subjects self-reporting their own behaviors, the reliability of annotation is not perfect. In some cases, this was the result of the subject reporting labels some time after the activity had occurred and mis-remembering the exact time. More common are cases where the subject neglected to report labels when relevant activities occurred (perhaps because the subject was distracted, did not have time to specify all the relevant labels, or was not aware of another relevant label in the vocabulary). As part of cleaning the data, we created adjusted versions for some labels using two methods: based on location data and based on other labels.

\emph{Location adjusted labels.} We collected absolute location coordinates of the examples that had location measurements (selecting the location update with best horizontal accuracy from each example) and visualized them on a map. This made it easier to correct some labels which were clearly reported incorrectly. In examples without location data the original label was maintained.
\begin{itemize}
	\item ``At the beach''. According to the few examples that reported being at the beach we marked areas that should be regarded as beach (and manually verified their validity by viewing them on a map). We then adjusted the label by applying ``At the beach'' to any example with a location within these areas.
	\item ``At home''. For each subject we identified the location of their home (by visualizing on a map all locations of examples where the subject reported being at home) and marked the coordinates of a visual centroid. This was only done when it was clear that we had indeed identified a location of a home. Three subjects reported being at home in two different houses, in which case we marked the two locations as locations of home. Two subjects never reported being at home but it was clear from their location to identify their location of residence. Some subjects had none or very few examples of ``at home'' with location data, so no home location was noted and their original reported labels were used. To define the adjusted version of the label ``at home'', whenever a subject's location was within 15 meters of their marked home location (or either of the two marked home locations), the adjusted value was set to ``true''; whenever a subject's location was farther than 100 meters from all the subject's marked home locations the adjusted value was set to ``false''. In other cases (when the location was between 15 and 100 meters from a home location, or when there was no location data available) we retained the subject's originally reported value for ``at home''. This adjustment removed some obviously false reports of ``at home'' (\eg, when the subject was clearly on a drive on a freeway). The adjustment manifested mostly by adding the missing label ``at home'' to many examples where the subject was clearly at home but failed to report it.
	\item ``At main workplace''. Similarly to home label we identified for each subject (if they used the original label ``at work'') the centroid location of their main workplace and created a new label --- ``at main workplace'' --- in a similar way. Some subjects reported being at work in different locations, so the original label ``at work'' is still valid for analysis and may have a different meaning than ``at main workplace'' (which was designed to capture behavioral patterns typical to the most common place that a person works in). This adjustment removed some examples where the label ``At work'' was probably incorrectly reported, but more importantly, it added the missing label in cases where the subject was clearly present at their most common workplace.
\end{itemize}

\emph{Labels corrected using other labels.} We used reported values of other labels to adjust some labels. In a few cases it was clear that the reported labels were mistakes (because the combination of labels was unreasonable). In other cases a relevant label was simply not reported, even though it  clearly should be relevant according to the other reported labels.
\begin{itemize}
	\item ``Walking''. We identified a few cases where subjects reported walking together with labels related to driving. In cases where location data was available, it was clear on the map that the correct activity was the drive and not the walk. In the adjusted version of ``walking'' we changed the value to ``false'' whenever the subject reported ``on a bus'', ``in a car'', ``drive (I'm the driver)'', ``drive (I'm a passenger)'', ``motorbike'', ``skateboarding'' or ``at the pool''.
	\item ``Running''. The adjusted version was set to ``false'' for the same activities as in the adjusted ``walking'' label, plus in cases where the subject reported ``playing baseball'' or ``playing frisbee''. Although these cases are likely still valid (because the subject decided to report they were running during these playing activities), we decided to create the adjusted ``running'' version to represent a more coherent running activity (assuming that the playing activities involve a mixture of running, walking and standing intermittently). While the adjusted versions of ``walking'' and ``running'' may have a few misses (\eg, some minutes during a baseball game when the subject was purely running), these misses don't harm the integrity of the multi-class experiments, which used only examples that had positive labels of ``running'', ``walking'', ``sitting'', \etc.
	\item ``Exercise''. The adjusted version was set to ``true'' whenever the subject reported ``exercising'', ``running'', ``bicycling'', ``lifting weights'', ``elliptical machine'', ``treadmill'', ``stationary bike'' or ``at the gym''. This adjustment boosted the representation of the exercise behavior and also took advantage of reported specific activities without enough examples to be analyzed on their own.
	\item ``Indoors''. The adjusted version was set to ``true'' whenever the subject reported ``indoors'', ``sleeping'', ``toilet'', ``bathing --- bath'', ``bathing --- shower'', ``in class'', ``at home'', ``at a bar'', ``at the gym'' or ``elevator''. It is reasonable that many subjects simply did not bother to report being indoors every time they did an activity indoors.
	\item ``Outside''. The adjusted version was set to ``true'' whenever the subject reported ``outside'', ``skateboarding'', ``playing baseball'', ``playing frisbee'', ``gardening'', ``raking leaves'', ``strolling'', ``hiking'', ``at the beach'', ``at sea'' or ``motorbike''.
	\item ``At a restaurant''. In the adjusted version we changed the value to ``false'' whenever the subject reported ``on a bus'', ``in a car'', ``drive (I'm the driver)'', ``drive (I'm a passenger)'' or ``motorbike''.
\end{itemize}

\subsection*{Classification methods}
Our system uses binary logistic regression classifiers (with a fitted intercept). Logistic regression provides a real-valued output, interpreted as the probability of the relevance of the label (value larger than 0.5 yielding a decision of ``relevant''). For each context label we used an  independent model.
We first randomly partitioned the training examples into internal training and validation subsets, allocating one third of the training examples for the validation subset, while maintaining the same proportion of positive \vs~negative examples in both subsets. We then used grid search to select the cost parameter $C$ for logistic regression: for each value (out of $\{0.001,0.01,0.1,1,10,100\}$) we trained a logistic regression model on the internal train subset and tested the model on the validation subset. We selected the value of $C$ that resulted in highest F1 measure on the validation subset. We then re-trained a logistic regression model with the selected value on the entire training set.
For the leave-one-user-out experiment with the EF system we simplified the procedure and only trained the logistic regression models with value of $C=1$ (instead of performing grid search).
The learned weights from LFL for a set of selected labels that are presented in Figure~\ref{fig:weightsAndConfmats}~(A) are taken from the first (of five) training set of the cross validation evaluation.
To look at misclassifications and to produce the confusion matrices in Figure~\ref{fig:weightsAndConfmats}~(B--G) we used the multiclass (one-versus-rest) version of logistic regression, with a fixed cost value of $C=1$. Each multiclass experiment used the set of examples annotated with exactly one label from the examined label subset and with data from all of the sensors of interest (so an experiment with only accelerometer and gyroscope sensors might have more examples than an experiment with accelerometer, gyroscope and watch accelerometer).
These experiments were more fitting than binary classification in cases where missing labels are common. For example, labels describing the phone's position were not always consistently annotated. A binary classifier will use all negative examples to learn a decision boundary, including examples the subject forgot to label, which may skew the results if there are many missing labels.

\subsection*{Performance evaluation}
In order to make a fair comparison among the different sensors, evaluation was done on the subset of examples with data from all six core sensors available ($\sim$177k examples from 51 subjects). In the training phase, however, a single-sensor classifier was allowed to use all examples available (\eg,~all examples in the dataset had phone state data, so the PS single-sensor classifier was trained with all examples). While the early fusion system benefited from the advantage of modeling correlations between features of different sensors, it was limited to being trained only on examples with all sensor data available. The late fusion systems, on the other hand, had the advantage of using single-sensor classifiers that were trained on many more examples.

Classifier performance was evaluated using 5-fold cross validation. The subjects were randomly partitioned once into 5 folds, while equalizing the proportion of iPhone \vs~Android users between folds (To keep a fair evaluation it was important to partition the subjects, and not randomly partition the pool of examples, in order to avoid having examples from the same subject appear in both the training set and the test set). The cross validation procedure repeats the following for each fold: (1) hold out the selected fold to act as the test set (2) train a classifier on the remaining folds (3) apply the classifier to the held out test set.
For each fold and for each label, we counted the numbers of true positives (TP\@. Examples that were correctly classified as positive), true negatives (TN\@. Examples that were correctly classified as negative), false positives (FP\@. Examples that were wrongfully classified as positive) and false negatives (FN\@. Examples that were wrongfully classified as negative). At the end of the 5-fold procedure we summed up the total numbers of TP, TN, FP and FN over the entire evaluation set and calculated the following performance metrics:
\begin{itemize}
	\item \emph{Accuracy} is the proportion of correctly classified examples out of all the examples. This metric is sensitive to imbalanced label proportion in the data.
	\item \emph{True positive rate} (TPR, also called sensitivity or recall) is the proportion of positive examples that were correctly classified as positive: recall=TPR=TP/(TP+FN). 
	\item \emph{True negative rate} (TNR, also called specificity) is the proportion of negative examples that were correctly classified as negative: TNR=TN/(TN+FP). 
	\item \emph{Precision} (prec) is the proportion of correctly classified examples out of the examples that the classifier declared as positive: precision=TP/(TP+FP). 
	\item \emph{Balanced accuracy} is a measure that accounts for the tradeoff between true positive rate and true negative rate: BA=(TPR+TNR)/2.
	\item The \emph{F1} measure is another such measure, which takes the harmonic mean of precision and recall: F1=(2*TPR*prec)/(TPR+prec). 
\end{itemize}

While the balanced accuracy is easy to interpret (chance level is always 0.5 and perfect performance is 1) the F1 measure is very sensitive to how rare the positive examples are, so for each label a typical F1 value is different.
The 5-fold subject partition is available with the dataset, and we encourage researchers using the dataset to evaluate new methods to use the same 5-fold partition, in order to promote fair comparisons.

\textbf{Random chance.}
To assess the statistical significance of the performance scores we achieved, we evaluated a distribution of performance scores achieved by a random classifier. The random classifier declares ``relevant'' with probability 0.5 independently for each example and for each label. To estimate the distribution of scores that such a classifier obtains, we ran 100 simulations (each time the classifier randomly assigned binary predictions and the performance scores were calculated over the entire evaluation dataset). Chance level (expected value of the random classifier) of balanced accuracy is 0.5 for every label. For the F1 measure the chance level for each label is dependent on the proportion of positive and negative examples in the data. For each performance measure and for each label we estimated a value which we call \emph{p99}, the 99$^{th}$ percentile among the 100 scores achieved in the 100 simulations. 
Hence the probability of obtaining a score greater than p99 by chance is less than 1\%. For average (over a set of labels) scores the p99 value was calculated similarly (computing the average score for each of the 100 simulations).

\subsection*{User personalization assessment}
To assess the advantages of user personalization, we selected a single test subject that had provided relatively many examples and many labels. We partitioned this user's examples into the first half and second half of the examples (according to their recording timestamps), to simulate an adaptation training period (the first half) and a deployment period (the second half). We used early fusion (EF) classifiers to combine the features from all 6 sensors.
The \emph{universal model} was the one used in previous experiments, taken from the fold where the test user was part of the cross validation test set (so the universal model was trained on 48 other users).
The \emph{individual model} was trained only on data from the test user, taken from the first half of the subject's examples.
The \emph{adapted model} was a combination of both the universal and individual models using the LFA method (\ie, averaging the probability outputs of both models).
All three models were tested on the same set of unseen test examples (the second half of the subject's examples).
For some labels, an insufficient number of examples to train an individual classifier resulted in a trivial classifier (always declaring the same answer). In those cases the performance was reported as chance level (BA of 0.5 and F1 of 0).

\clearpage
\section*{Detailed results tables}
\subsection*{5-fold cross validation evaluation}

\begin{table*}[h]
	\centering
	\hspace*{-1cm}
	\begin{tabular}{l|c|c|c||c|c|c|c|c|c||c|c|c}
		& $n_e$ & $n_s$ & p99& \textbf{Acc}& \textbf{Gyro}& \textbf{WAcc}& \textbf{Loc}& \textbf{Aud}& \textbf{PS}& \textbf{EF}& \textbf{LFA}& \textbf{LFL} \\
		\hline
		Lying down & 54359 & 47 & 0.50  & 0.72 & 0.69 & 0.81 & 0.66 & 0.79 & 0.85 & 0.87 & 0.86 & \textbf{0.88} \\
		Sitting & 82904 & 50 & 0.50  & 0.63 & 0.61 & 0.68 & 0.61 & 0.65 & 0.69 & \textbf{0.76} & 0.75 & 0.75 \\
		Walking & 11892 & 50 & 0.51  & 0.77 & 0.80 & 0.75 & 0.66 & 0.63 & 0.70 & 0.80 & 0.80 & \textbf{0.80} \\
		Running & 675 & 19 & 0.52  & 0.69 & 0.66 & \textbf{0.80} & 0.56 & 0.48 & 0.58 & 0.62 & 0.69 & 0.71 \\
		Bicycling & 3523 & 22 & 0.51  & 0.81 & 0.81 & 0.84 & 0.81 & 0.77 & 0.83 & 0.87 & 0.87 & \textbf{0.87} \\
		Sleeping & 42920 & 40 & 0.50  & 0.75 & 0.70 & 0.81 & 0.62 & 0.79 & 0.87 & 0.88 & 0.88 & \textbf{0.89} \\
		Lab work & 2898 & 8 & 0.51  & 0.71 & 0.62 & 0.65 & 0.84 & 0.71 & 0.81 & 0.84 & \textbf{0.87} & 0.85 \\
		In class & 2872 & 13 & 0.51  & 0.60 & 0.63 & 0.57 & 0.74 & 0.76 & 0.67 & 0.70 & 0.77 & \textbf{0.80} \\
		In a meeting & 2904 & 34 & 0.51  & 0.60 & 0.57 & 0.62 & 0.63 & 0.79 & 0.73 & 0.80 & 0.79 & \textbf{0.82} \\
		At main workplace & 20382 & 26 & 0.50  & 0.57 & 0.49 & 0.63 & 0.76 & 0.65 & 0.78 & 0.80 & 0.80 & \textbf{0.81} \\
		Indoors & 107944 & 51 & 0.50  & 0.66 & 0.66 & 0.67 & 0.63 & 0.71 & 0.72 & 0.75 & 0.75 & \textbf{0.76} \\
		Outside & 7629 & 36 & 0.51  & 0.70 & 0.73 & 0.70 & 0.66 & 0.66 & 0.73 & 0.74 & 0.77 & \textbf{0.78} \\
		In a car & 3635 & 24 & 0.51  & 0.79 & 0.65 & 0.71 & 0.81 & 0.77 & 0.84 & 0.85 & 0.86 & \textbf{0.86} \\
		On a bus & 1185 & 24 & 0.52  & 0.73 & 0.69 & 0.67 & 0.75 & 0.74 & 0.82 & 0.77 & \textbf{0.84} & 0.83 \\
		Drive (I'm the driver) & 5034 & 24 & 0.51  & 0.79 & 0.61 & 0.75 & 0.82 & 0.74 & 0.83 & 0.84 & 0.86 & \textbf{0.87} \\
		Drive (I'm a passenger) & 1655 & 19 & 0.51  & 0.76 & 0.71 & 0.64 & 0.79 & 0.76 & 0.81 & 0.84 & 0.84 & \textbf{0.85} \\
		At home & 83977 & 50 & 0.50  & 0.65 & 0.63 & 0.66 & 0.63 & 0.71 & 0.70 & 0.75 & 0.77 & \textbf{0.78} \\
		At a restaurant & 1320 & 16 & 0.52  & 0.62 & 0.67 & 0.68 & 0.58 & \textbf{0.85} & 0.77 & 0.76 & 0.83 & 0.81 \\
		Phone in pocket & 15301 & 31 & 0.50  & 0.69 & 0.75 & 0.67 & 0.61 & 0.64 & 0.72 & 0.77 & \textbf{0.77} & 0.77 \\
		Exercise & 5384 & 36 & 0.51  & 0.73 & 0.73 & 0.77 & 0.71 & 0.70 & 0.77 & 0.81 & 0.80 & \textbf{0.81} \\
		Cooking & 2257 & 33 & 0.51  & 0.52 & 0.53 & 0.68 & 0.57 & 0.62 & 0.68 & 0.71 & 0.71 & \textbf{0.72} \\
		Shopping & 896 & 18 & 0.52  & 0.70 & 0.70 & 0.69 & 0.54 & 0.59 & 0.79 & 0.69 & 0.76 & \textbf{0.80} \\
		Strolling & 434 & 8 & 0.53  & 0.67 & 0.74 & 0.72 & 0.67 & 0.64 & 0.75 & 0.66 & \textbf{0.77} & 0.74 \\
		Drinking (alcohol) & 864 & 10 & 0.52  & 0.71 & 0.69 & 0.50 & 0.56 & 0.80 & 0.74 & 0.70 & \textbf{0.82} & 0.81 \\
		Bathing - shower & 1186 & 27 & 0.52  & 0.53 & 0.55 & \textbf{0.73} & 0.47 & 0.63 & 0.47 & 0.64 & 0.67 & 0.70 \\
		\hline
		average & &  & 0.50  & 0.68 & 0.66 & 0.70 & 0.67 & 0.70 & 0.75 & 0.77 & 0.80 & 0.80 \\
	\end{tabular}
	\caption{5-fold evaluation performance (BA) of the different classifiers on each label. Part 1 of the labels. For each label $n_e$ is the number of examples and $n_s$ is the number of subjects in the testing (possibly more examples participated in the training). p99 marks the 99$^{th}$ percentile of random scores --- a score above the p99 value has less than 0.01 probability to be achieved randomly. For each label the score of the highest performing classifier is marked in bold.}
\label{tab:atrPerLabel1.5fold}
\end{table*}

\begin{table*}[h]
	\centering
	\hspace*{-1cm}
	\begin{tabular}{l|c|c|c||c|c|c|c|c|c||c|c|c}
		& $n_e$ & $n_s$ & p99& \textbf{Acc}& \textbf{Gyro}& \textbf{WAcc}& \textbf{Loc}& \textbf{Aud}& \textbf{PS}& \textbf{EF}& \textbf{LFA}& \textbf{LFL} \\
		\hline
		Cleaning & 1839 & 22 & 0.51  & 0.63 & 0.64 & \textbf{0.71} & 0.41 & 0.60 & 0.51 & 0.60 & 0.70 & 0.68 \\
		Laundry & 473 & 12 & 0.52  & 0.65 & 0.66 & \textbf{0.66} & 0.38 & 0.53 & 0.65 & 0.58 & 0.63 & 0.63 \\
		Washing dishes & 851 & 17 & 0.52  & 0.40 & 0.52 & 0.70 & 0.58 & 0.60 & 0.57 & 0.65 & 0.70 & \textbf{0.70} \\
		Watching TV & 9412 & 28 & 0.51  & 0.61 & 0.54 & 0.56 & 0.56 & 0.64 & 0.67 & 0.65 & \textbf{0.70} & 0.68 \\
		Surfing the internet & 11641 & 28 & 0.50  & 0.56 & 0.55 & 0.60 & 0.54 & 0.60 & 0.57 & 0.59 & 0.63 & \textbf{0.63} \\
		At a party & 404 & 3 & 0.53  & 0.74 & 0.71 & 0.49 & 0.54 & \textbf{0.81} & 0.56 & 0.54 & 0.76 & 0.75 \\
		At a bar & 520 & 4 & 0.53  & 0.45 & 0.66 & 0.53 & 0.60 & 0.49 & \textbf{0.93} & 0.50 & 0.61 & 0.66 \\
		At the beach & 122 & 5 & 0.55  & 0.62 & 0.48 & 0.47 & \textbf{0.72} & 0.58 & 0.70 & 0.50 & 0.71 & 0.70 \\
		Singing & 384 & 6 & 0.53  & 0.57 & 0.64 & 0.46 & 0.61 & \textbf{0.68} & 0.59 & 0.50 & 0.65 & 0.53 \\
		Talking & 18976 & 44 & 0.50  & 0.60 & 0.61 & 0.60 & 0.54 & 0.65 & 0.65 & 0.65 & 0.67 & \textbf{0.67} \\
		Computer work & 23692 & 38 & 0.50  & 0.57 & 0.56 & 0.62 & 0.63 & 0.61 & 0.68 & 0.68 & \textbf{0.71} & 0.70 \\
		Eating & 10169 & 49 & 0.51  & 0.59 & 0.58 & 0.60 & 0.51 & 0.61 & 0.62 & \textbf{0.66} & 0.65 & 0.65 \\
		Toilet & 1646 & 33 & 0.51  & 0.57 & 0.51 & 0.58 & 0.57 & 0.64 & 0.59 & 0.65 & 0.66 & \textbf{0.66} \\
		Grooming & 1847 & 25 & 0.51  & 0.44 & 0.49 & 0.62 & 0.59 & 0.63 & 0.58 & 0.60 & \textbf{0.63} & 0.63 \\
		Dressing & 1308 & 27 & 0.52  & 0.51 & 0.52 & 0.64 & 0.54 & 0.65 & 0.61 & 0.64 & \textbf{0.67} & 0.67 \\
		At the gym & 906 & 6 & 0.52  & 0.50 & 0.55 & 0.58 & 0.57 & 0.65 & \textbf{0.70} & 0.54 & 0.64 & 0.61 \\
		Stairs - going up & 399 & 17 & 0.53  & 0.70 & \textbf{0.73} & 0.65 & 0.55 & 0.55 & 0.51 & 0.58 & 0.69 & 0.67 \\
		Stairs - going down & 390 & 15 & 0.53  & 0.71 & \textbf{0.73} & 0.66 & 0.55 & 0.55 & 0.51 & 0.58 & 0.71 & 0.66 \\
		Elevator & 124 & 8 & 0.55  & 0.72 & \textbf{0.76} & 0.44 & 0.54 & 0.71 & 0.51 & 0.49 & 0.73 & 0.73 \\
		Standing & 22766 & 51 & 0.50  & 0.60 & 0.59 & 0.67 & 0.54 & 0.59 & 0.63 & \textbf{0.68} & 0.67 & 0.68 \\
		At school & 25840 & 39 & 0.50  & 0.59 & 0.59 & 0.59 & 0.66 & 0.64 & 0.68 & 0.70 & \textbf{0.70} & 0.70 \\
		Phone in hand & 8595 & 37 & 0.51  & 0.65 & \textbf{0.68} & 0.56 & 0.59 & 0.59 & 0.61 & 0.64 & 0.67 & 0.66 \\
		Phone in bag & 5589 & 22 & 0.51  & 0.59 & 0.56 & 0.55 & 0.59 & 0.64 & 0.69 & 0.67 & 0.68 & \textbf{0.69} \\
		Phone on table & 70611 & 43 & 0.50  & 0.60 & 0.61 & 0.56 & 0.53 & 0.55 & 0.61 & 0.61 & 0.62 & \textbf{0.62} \\
		With co-workers & 4139 & 17 & 0.51  & 0.57 & 0.57 & 0.61 & 0.58 & 0.68 & 0.67 & 0.69 & 0.71 & \textbf{0.72} \\
		With friends & 12865 & 25 & 0.50  & 0.55 & 0.58 & 0.53 & 0.54 & 0.60 & 0.60 & 0.55 & \textbf{0.61} & 0.58 \\
		\hline
		average & &  & 0.50  & 0.59 & 0.60 & 0.59 & 0.56 & 0.62 & 0.62 & 0.60 & 0.67 & 0.66 \\
	\end{tabular}
	\caption{5-fold evaluation performance (BA) of the different classifiers on each label. Part 2 of the labels. For each label $n_e$ is the number of examples and $n_s$ is the number of subjects in the testing (possibly more examples participated in the training). p99 marks the 99$^{th}$ percentile of random scores --- a score above the p99 value has less than 0.01 probability to be achieved randomly. For each label the score of the highest performing classifier is marked in bold.}
\label{tab:atrPerLabel2.5fold}
\end{table*}

\begin{table*}[h]
	\centering
	\hspace*{-1cm}
	\begin{tabular}{l|c|c|c||c|c|c|c|c|c||c|c|c}
		& $n_e$ & $n_s$ & p99& \textbf{Acc}& \textbf{Gyro}& \textbf{WAcc}& \textbf{Loc}& \textbf{Aud}& \textbf{PS}& \textbf{EF}& \textbf{LFA}& \textbf{LFL} \\
		\hline
		Lying down & 54359 & 47 & 0.38  & 0.61 & 0.59 & 0.71 & 0.55 & 0.69 & 0.78 & 0.81 & 0.79 & \textbf{0.82} \\
		Sitting & 82904 & 50 & 0.49  & 0.58 & 0.58 & 0.67 & 0.59 & 0.62 & 0.72 & \textbf{0.75} & 0.75 & 0.74 \\
		Walking & 11892 & 50 & 0.12  & 0.38 & 0.38 & 0.31 & 0.22 & 0.19 & 0.22 & 0.38 & \textbf{0.39} & 0.38 \\
		Running & 675 & 19 & 0.01  & 0.03 & 0.03 & \textbf{0.04} & 0.01 & 0.01 & 0.01 & 0.03 & 0.04 & 0.04 \\
		Bicycling & 3523 & 22 & 0.04  & 0.19 & 0.16 & 0.23 & 0.18 & 0.12 & 0.17 & \textbf{0.31} & 0.25 & 0.26 \\
		Sleeping & 42920 & 40 & 0.33  & 0.57 & 0.53 & 0.65 & 0.44 & 0.63 & 0.75 & 0.79 & 0.77 & \textbf{0.81} \\
		Lab work & 2898 & 8 & 0.03  & 0.08 & 0.06 & 0.06 & 0.11 & 0.09 & 0.11 & \textbf{0.21} & 0.16 & 0.19 \\
		In class & 2872 & 13 & 0.03  & 0.05 & 0.06 & 0.04 & 0.07 & 0.10 & 0.06 & 0.13 & 0.12 & \textbf{0.14} \\
		In a meeting & 2904 & 34 & 0.03  & 0.05 & 0.04 & 0.05 & 0.05 & 0.11 & 0.07 & \textbf{0.17} & 0.10 & 0.14 \\
		At main workplace & 20382 & 26 & 0.19  & 0.23 & 0.18 & 0.28 & 0.41 & 0.31 & 0.42 & 0.49 & 0.47 & \textbf{0.50} \\
		Indoors & 107944 & 51 & 0.55  & 0.74 & 0.73 & 0.70 & 0.68 & 0.75 & 0.71 & 0.78 & \textbf{0.79} & 0.78 \\
		Outside & 7629 & 36 & 0.08  & 0.20 & 0.20 & 0.18 & 0.15 & 0.15 & 0.20 & 0.23 & \textbf{0.26} & 0.25 \\
		In a car & 3635 & 24 & 0.04  & 0.15 & 0.07 & 0.10 & \textbf{0.27} & 0.13 & 0.16 & 0.23 & 0.22 & 0.23 \\
		On a bus & 1185 & 24 & 0.01  & 0.04 & 0.03 & 0.03 & 0.07 & 0.04 & 0.06 & 0.07 & \textbf{0.07} & 0.06 \\
		Drive (I'm the driver) & 5034 & 24 & 0.06  & 0.21 & 0.09 & 0.15 & \textbf{0.37} & 0.16 & 0.23 & 0.31 & 0.31 & 0.31 \\
		Drive (I'm a passenger) & 1655 & 19 & 0.02  & 0.07 & 0.04 & 0.04 & \textbf{0.15} & 0.07 & 0.08 & 0.14 & 0.12 & 0.12 \\
		At home & 83977 & 50 & 0.49  & 0.66 & 0.65 & 0.64 & 0.63 & 0.70 & 0.67 & 0.74 & 0.76 & \textbf{0.77} \\
		At a restaurant & 1320 & 16 & 0.02  & 0.02 & 0.03 & 0.03 & 0.02 & 0.08 & 0.05 & \textbf{0.11} & 0.07 & 0.09 \\
		Phone in pocket & 15301 & 31 & 0.15  & 0.28 & 0.33 & 0.26 & 0.21 & 0.25 & 0.28 & \textbf{0.38} & 0.36 & 0.37 \\
		Exercise & 5384 & 36 & 0.06  & 0.21 & 0.18 & 0.24 & 0.14 & 0.13 & 0.19 & \textbf{0.27} & 0.26 & 0.25 \\
		Cooking & 2257 & 33 & 0.03  & 0.03 & 0.03 & 0.05 & 0.03 & 0.04 & 0.06 & \textbf{0.09} & 0.07 & 0.08 \\
		Shopping & 896 & 18 & 0.01  & 0.03 & 0.03 & 0.02 & 0.01 & 0.02 & 0.04 & \textbf{0.04} & 0.04 & 0.04 \\
		Strolling & 434 & 8 & 0.01  & 0.02 & 0.02 & 0.01 & 0.01 & 0.01 & 0.02 & 0.03 & \textbf{0.03} & 0.03 \\
		Drinking (alcohol) & 864 & 10 & 0.01  & 0.03 & 0.02 & 0.01 & 0.01 & 0.04 & 0.03 & \textbf{0.07} & 0.07 & 0.06 \\
		Bathing - shower & 1186 & 27 & 0.01  & 0.01 & 0.02 & 0.04 & 0.01 & 0.02 & 0.01 & 0.04 & 0.04 & \textbf{0.05} \\
		\hline
		average & &  & 0.13  & 0.22 & 0.20 & 0.22 & 0.22 & 0.22 & 0.24 & 0.30 & 0.29 & 0.30 \\
	\end{tabular}
	\caption{5-fold evaluation performance (F1) of the different classifiers on each label. Part 1 of the labels. For each label $n_e$ is the number of examples and $n_s$ is the number of subjects in the testing (possibly more examples participated in the training). p99 marks the 99$^{th}$ percentile of random scores --- a score above the p99 value has less than 0.01 probability to be achieved randomly. For each label the score of the highest performing classifier is marked in bold.}
\label{tab:f1PerLabel1.5fold}
\end{table*}

\begin{table*}[h]
	\centering
	\hspace*{-1cm}
	\begin{tabular}{l|c|c|c||c|c|c|c|c|c||c|c|c}
		& $n_e$ & $n_s$ & p99& \textbf{Acc}& \textbf{Gyro}& \textbf{WAcc}& \textbf{Loc}& \textbf{Aud}& \textbf{PS}& \textbf{EF}& \textbf{LFA}& \textbf{LFL} \\
		\hline
		Cleaning & 1839 & 22 & 0.02  & 0.05 & 0.05 & 0.05 & 0.01 & 0.03 & 0.02 & 0.05 & \textbf{0.06} & 0.05 \\
		Laundry & 473 & 12 & 0.01  & 0.01 & 0.01 & 0.01 & 0.00 & 0.01 & 0.01 & \textbf{0.02} & 0.01 & 0.02 \\
		Washing dishes & 851 & 17 & 0.01  & 0.01 & 0.01 & 0.03 & 0.01 & 0.02 & 0.01 & 0.03 & 0.03 & \textbf{0.04} \\
		Watching TV & 9412 & 28 & 0.10  & 0.14 & 0.11 & 0.12 & 0.12 & 0.17 & 0.18 & 0.21 & 0.22 & \textbf{0.22} \\
		Surfing the internet & 11641 & 28 & 0.12  & 0.15 & 0.14 & 0.17 & 0.13 & 0.17 & 0.15 & 0.18 & 0.19 & \textbf{0.20} \\
		At a party & 404 & 3 & 0.01  & 0.01 & 0.01 & 0.00 & 0.01 & 0.03 & 0.01 & \textbf{0.04} & 0.03 & 0.04 \\
		At a bar & 520 & 4 & 0.01  & 0.00 & 0.01 & 0.01 & 0.01 & 0.00 & \textbf{0.09} & 0.00 & 0.03 & 0.06 \\
		At the beach & 122 & 5 & 0.00  & 0.00 & 0.00 & 0.00 & 0.01 & 0.00 & 0.01 & 0.00 & \textbf{0.02} & 0.02 \\
		Singing & 384 & 6 & 0.00  & 0.01 & 0.01 & 0.00 & 0.01 & 0.01 & 0.01 & 0.00 & \textbf{0.01} & 0.01 \\
		Talking & 18976 & 44 & 0.18  & 0.25 & 0.26 & 0.24 & 0.21 & 0.29 & 0.27 & 0.29 & 0.30 & \textbf{0.30} \\
		Computer work & 23692 & 38 & 0.21  & 0.26 & 0.25 & 0.30 & 0.31 & 0.30 & 0.35 & 0.38 & 0.39 & \textbf{0.39} \\
		Eating & 10169 & 49 & 0.11  & 0.15 & 0.14 & 0.15 & 0.11 & 0.16 & 0.15 & \textbf{0.19} & 0.18 & 0.17 \\
		Toilet & 1646 & 33 & 0.02  & 0.03 & 0.02 & 0.03 & 0.02 & 0.03 & 0.03 & \textbf{0.04} & 0.04 & 0.04 \\
		Grooming & 1847 & 25 & 0.02  & 0.02 & 0.02 & 0.03 & 0.03 & 0.04 & 0.03 & 0.05 & 0.04 & \textbf{0.05} \\
		Dressing & 1308 & 27 & 0.02  & 0.02 & 0.02 & 0.03 & 0.02 & 0.03 & 0.03 & \textbf{0.04} & 0.04 & 0.04 \\
		At the gym & 906 & 6 & 0.01  & 0.01 & 0.01 & 0.02 & 0.01 & 0.02 & 0.02 & 0.03 & 0.03 & \textbf{0.03} \\
		Stairs - going up & 399 & 17 & 0.01  & \textbf{0.02} & 0.02 & 0.01 & 0.01 & 0.01 & 0.00 & 0.01 & 0.02 & 0.01 \\
		Stairs - going down & 390 & 15 & 0.00  & \textbf{0.02} & 0.02 & 0.01 & 0.01 & 0.01 & 0.00 & 0.01 & 0.02 & 0.01 \\
		Elevator & 124 & 8 & 0.00  & 0.01 & 0.01 & 0.00 & 0.00 & 0.00 & 0.00 & 0.00 & 0.01 & \textbf{0.01} \\
		Standing & 22766 & 51 & 0.21  & 0.28 & 0.27 & 0.35 & 0.23 & 0.27 & 0.29 & \textbf{0.36} & 0.34 & 0.35 \\
		At school & 25840 & 39 & 0.23  & 0.30 & 0.29 & 0.30 & 0.38 & 0.34 & 0.39 & 0.41 & 0.41 & \textbf{0.41} \\
		Phone in hand & 8595 & 37 & 0.09  & 0.17 & \textbf{0.17} & 0.11 & 0.13 & 0.13 & 0.13 & 0.16 & 0.17 & 0.16 \\
		Phone in bag & 5589 & 22 & 0.06  & 0.10 & 0.08 & 0.07 & 0.09 & 0.11 & 0.12 & \textbf{0.15} & 0.14 & 0.14 \\
		Phone on table & 70611 & 43 & 0.45  & 0.58 & 0.58 & 0.51 & 0.50 & 0.51 & 0.56 & 0.56 & \textbf{0.59} & 0.58 \\
		With co-workers & 4139 & 17 & 0.05  & 0.06 & 0.06 & 0.07 & 0.06 & 0.11 & 0.08 & 0.13 & 0.12 & \textbf{0.13} \\
		With friends & 12865 & 25 & 0.13  & 0.15 & 0.17 & 0.14 & 0.15 & 0.18 & 0.18 & 0.15 & \textbf{0.19} & 0.18 \\
		\hline
		average & &  & 0.08  & 0.11 & 0.11 & 0.11 & 0.10 & 0.11 & 0.12 & 0.13 & 0.14 & 0.14 \\
	\end{tabular}
	\caption{5-fold evaluation performance (F1) of the different classifiers on each label. Part 2 of the labels. For each label $n_e$ is the number of examples and $n_s$ is the number of subjects in the testing (possibly more examples participated in the training). p99 marks the 99$^{th}$ percentile of random scores --- a score above the p99 value has less than 0.01 probability to be achieved randomly. For each label the score of the highest performing classifier is marked in bold.}
\label{tab:f1PerLabel2.5fold}
\end{table*}

\clearpage

\subsection*{Leave-one-user-out evaluation}

\begin{table*}[h]
	\centering
	\hspace*{-1cm}
	\begin{tabular}{l|c|c|c||c|c|c|c|c|c||c|c|c}
		& $n_e$ & $n_s$ & p99& \textbf{Acc}& \textbf{Gyro}& \textbf{WAcc}& \textbf{Loc}& \textbf{Aud}& \textbf{PS}& \textbf{EF}& \textbf{LFA}& \textbf{LFL} \\
		\hline
		Lying down & 54359 & 47 & 0.50  & 0.73 & 0.69 & 0.81 & 0.65 & 0.79 & 0.84 & 0.87 & 0.86 & \textbf{0.88} \\
		Sitting & 82904 & 50 & 0.50  & 0.63 & 0.61 & 0.68 & 0.61 & 0.65 & 0.69 & \textbf{0.76} & 0.75 & 0.75 \\
		Walking & 11892 & 50 & 0.51  & 0.77 & 0.80 & 0.75 & 0.66 & 0.63 & 0.71 & 0.80 & 0.80 & \textbf{0.81} \\
		Running & 675 & 19 & 0.52  & 0.69 & 0.69 & \textbf{0.80} & 0.56 & 0.50 & 0.57 & 0.67 & 0.72 & 0.76 \\
		Bicycling & 3523 & 22 & 0.51  & 0.81 & 0.81 & 0.85 & 0.80 & 0.76 & 0.80 & \textbf{0.87} & 0.86 & 0.87 \\
		Sleeping & 42920 & 40 & 0.50  & 0.75 & 0.70 & 0.81 & 0.62 & 0.80 & 0.85 & 0.88 & 0.88 & \textbf{0.89} \\
		Lab work & 2898 & 8 & 0.51  & 0.69 & 0.61 & 0.65 & 0.82 & 0.70 & 0.84 & 0.84 & \textbf{0.84} & 0.84 \\
		In class & 2872 & 13 & 0.51  & 0.61 & 0.63 & 0.58 & 0.74 & 0.77 & 0.72 & 0.74 & 0.79 & \textbf{0.81} \\
		In a meeting & 2904 & 34 & 0.51  & 0.62 & 0.59 & 0.62 & 0.66 & 0.78 & 0.73 & 0.81 & 0.80 & \textbf{0.82} \\
		At main workplace & 20382 & 26 & 0.50  & 0.55 & 0.49 & 0.64 & 0.76 & 0.65 & 0.80 & 0.80 & 0.81 & \textbf{0.82} \\
		Indoors & 107944 & 51 & 0.50  & 0.67 & 0.66 & 0.68 & 0.63 & 0.70 & 0.72 & \textbf{0.76} & 0.75 & 0.76 \\
		Outside & 7629 & 36 & 0.51  & 0.72 & 0.74 & 0.70 & 0.66 & 0.67 & 0.71 & 0.75 & 0.78 & \textbf{0.79} \\
		In a car & 3635 & 24 & 0.51  & 0.79 & 0.66 & 0.71 & 0.82 & 0.76 & 0.83 & 0.85 & 0.86 & \textbf{0.87} \\
		On a bus & 1185 & 24 & 0.52  & 0.74 & 0.69 & 0.68 & 0.72 & 0.72 & 0.80 & 0.78 & \textbf{0.83} & 0.82 \\
		Drive (I'm the driver) & 5034 & 24 & 0.51  & 0.80 & 0.62 & 0.75 & 0.83 & 0.75 & 0.84 & 0.84 & 0.86 & \textbf{0.87} \\
		Drive (I'm a passenger) & 1655 & 19 & 0.51  & 0.76 & 0.70 & 0.64 & 0.80 & 0.77 & 0.82 & \textbf{0.84} & 0.83 & 0.84 \\
		At home & 83977 & 50 & 0.50  & 0.65 & 0.63 & 0.66 & 0.62 & 0.72 & 0.72 & 0.76 & 0.77 & \textbf{0.77} \\
		At a restaurant & 1320 & 16 & 0.52  & 0.62 & 0.68 & 0.69 & 0.57 & \textbf{0.84} & 0.74 & 0.79 & 0.84 & 0.84 \\
		Phone in pocket & 15301 & 31 & 0.50  & 0.69 & 0.75 & 0.67 & 0.61 & 0.64 & 0.71 & 0.77 & 0.77 & \textbf{0.77} \\
		Exercise & 5384 & 36 & 0.51  & 0.74 & 0.73 & 0.77 & 0.71 & 0.70 & 0.75 & 0.81 & 0.81 & \textbf{0.81} \\
		Cooking & 2257 & 33 & 0.51  & 0.52 & 0.55 & 0.67 & 0.57 & 0.62 & 0.68 & 0.71 & 0.72 & \textbf{0.72} \\
		Shopping & 896 & 18 & 0.52  & 0.71 & 0.69 & 0.68 & 0.53 & 0.57 & \textbf{0.79} & 0.67 & 0.75 & 0.78 \\
		Strolling & 434 & 8 & 0.53  & 0.64 & 0.73 & 0.70 & 0.63 & 0.62 & 0.71 & 0.67 & 0.74 & \textbf{0.75} \\
		Drinking (alcohol) & 864 & 10 & 0.52  & 0.72 & 0.70 & 0.54 & 0.56 & 0.79 & 0.54 & 0.68 & \textbf{0.79} & 0.77 \\
		Bathing - shower & 1186 & 27 & 0.52  & 0.50 & 0.54 & \textbf{0.74} & 0.48 & 0.63 & 0.48 & 0.64 & 0.69 & 0.72 \\
		\hline
		average & &  & 0.50  & 0.68 & 0.67 & 0.70 & 0.66 & 0.70 & 0.74 & 0.78 & 0.80 & 0.81 \\
	\end{tabular}
	\caption{Leave-one-user-out evaluation performance (BA) of the different classifiers on each label. Part 1 of the labels. For each label $n_e$ is the number of examples and $n_s$ is the number of subjects in the testing (possibly more examples participated in the training). p99 marks the 99$^{th}$ percentile of random scores --- a score above the p99 value has less than 0.01 probability to be achieved randomly. For each label the score of the highest performing classifier is marked in bold.}
\label{tab:atrPerLabel1.loo}
\end{table*}

\begin{table*}[h]
	\centering
	\hspace*{-1cm}
	\begin{tabular}{l|c|c|c||c|c|c|c|c|c||c|c|c}
		& $n_e$ & $n_s$ & p99& \textbf{Acc}& \textbf{Gyro}& \textbf{WAcc}& \textbf{Loc}& \textbf{Aud}& \textbf{PS}& \textbf{EF}& \textbf{LFA}& \textbf{LFL} \\
		\hline
		Cleaning & 1839 & 22 & 0.51  & 0.62 & 0.63 & \textbf{0.73} & 0.42 & 0.62 & 0.49 & 0.70 & 0.71 & 0.70 \\
		Laundry & 473 & 12 & 0.52  & 0.67 & 0.65 & 0.66 & 0.35 & 0.52 & \textbf{0.78} & 0.60 & 0.68 & 0.70 \\
		Washing dishes & 851 & 17 & 0.52  & 0.36 & 0.48 & \textbf{0.69} & 0.54 & 0.61 & 0.54 & 0.66 & 0.67 & 0.68 \\
		Watching TV & 9412 & 28 & 0.51  & 0.61 & 0.54 & 0.57 & 0.57 & 0.67 & 0.66 & 0.69 & \textbf{0.72} & 0.71 \\
		Surfing the internet & 11641 & 28 & 0.50  & 0.55 & 0.58 & 0.59 & 0.56 & 0.60 & 0.57 & 0.61 & \textbf{0.62} & 0.62 \\
		At a party & 404 & 3 & 0.53  & 0.73 & 0.71 & 0.48 & 0.70 & \textbf{0.84} & 0.67 & 0.52 & 0.79 & 0.76 \\
		At a bar & 520 & 4 & 0.53  & 0.53 & 0.69 & 0.50 & 0.64 & 0.62 & \textbf{0.88} & 0.52 & 0.71 & 0.68 \\
		At the beach & 122 & 5 & 0.55  & 0.66 & 0.51 & 0.52 & \textbf{0.71} & 0.58 & 0.69 & 0.57 & 0.71 & 0.71 \\
		Singing & 384 & 6 & 0.53  & 0.56 & 0.62 & 0.46 & \textbf{0.70} & 0.68 & 0.60 & 0.48 & 0.68 & 0.53 \\
		Talking & 18976 & 44 & 0.50  & 0.61 & 0.61 & 0.61 & 0.55 & 0.66 & 0.64 & 0.66 & 0.68 & \textbf{0.68} \\
		Computer work & 23692 & 38 & 0.50  & 0.59 & 0.57 & 0.62 & 0.65 & 0.59 & 0.67 & 0.69 & \textbf{0.71} & 0.69 \\
		Eating & 10169 & 49 & 0.51  & 0.59 & 0.58 & 0.60 & 0.53 & 0.61 & 0.63 & 0.66 & \textbf{0.66} & 0.66 \\
		Toilet & 1646 & 33 & 0.51  & 0.57 & 0.52 & 0.57 & 0.57 & 0.63 & 0.56 & 0.65 & \textbf{0.65} & 0.65 \\
		Grooming & 1847 & 25 & 0.51  & 0.46 & 0.53 & 0.62 & 0.60 & 0.65 & 0.53 & 0.63 & 0.64 & \textbf{0.66} \\
		Dressing & 1308 & 27 & 0.52  & 0.51 & 0.54 & 0.66 & 0.53 & 0.67 & 0.55 & 0.66 & 0.67 & \textbf{0.68} \\
		At the gym & 906 & 6 & 0.52  & 0.55 & 0.56 & 0.67 & 0.51 & 0.67 & \textbf{0.70} & 0.58 & 0.67 & 0.67 \\
		Stairs - going up & 399 & 17 & 0.53  & 0.68 & \textbf{0.76} & 0.65 & 0.57 & 0.57 & 0.48 & 0.59 & 0.69 & 0.66 \\
		Stairs - going down & 390 & 15 & 0.53  & 0.70 & \textbf{0.75} & 0.66 & 0.54 & 0.55 & 0.48 & 0.57 & 0.69 & 0.63 \\
		Elevator & 124 & 8 & 0.55  & 0.68 & 0.70 & 0.56 & 0.57 & \textbf{0.70} & 0.54 & 0.50 & 0.62 & 0.61 \\
		Standing & 22766 & 51 & 0.50  & 0.60 & 0.59 & 0.67 & 0.54 & 0.59 & 0.62 & \textbf{0.68} & 0.66 & 0.67 \\
		At school & 25840 & 39 & 0.50  & 0.60 & 0.59 & 0.59 & 0.68 & 0.66 & 0.70 & \textbf{0.72} & 0.71 & 0.71 \\
		Phone in hand & 8595 & 37 & 0.51  & 0.66 & \textbf{0.68} & 0.56 & 0.58 & 0.58 & 0.59 & 0.63 & 0.67 & 0.66 \\
		Phone in bag & 5589 & 22 & 0.51  & 0.60 & 0.56 & 0.56 & 0.59 & 0.69 & 0.69 & 0.72 & 0.71 & \textbf{0.73} \\
		Phone on table & 70611 & 43 & 0.50  & 0.60 & 0.61 & 0.56 & 0.52 & 0.56 & 0.61 & 0.61 & \textbf{0.63} & 0.62 \\
		With co-workers & 4139 & 17 & 0.51  & 0.55 & 0.57 & 0.61 & 0.61 & 0.68 & 0.71 & 0.69 & 0.73 & \textbf{0.74} \\
		With friends & 12865 & 25 & 0.50  & 0.56 & 0.57 & 0.54 & 0.55 & 0.62 & 0.59 & 0.58 & \textbf{0.63} & 0.61 \\
		\hline
		average & &  & 0.50  & 0.59 & 0.60 & 0.60 & 0.57 & 0.63 & 0.62 & 0.62 & 0.68 & 0.67 \\
	\end{tabular}
	\caption{Leave-one-user-out evaluation performance (BA) of the different classifiers on each label. Part 2 of the labels. For each label $n_e$ is the number of examples and $n_s$ is the number of subjects in the testing (possibly more examples participated in the training). p99 marks the 99$^{th}$ percentile of random scores --- a score above the p99 value has less than 0.01 probability to be achieved randomly. For each label the score of the highest performing classifier is marked in bold.}
\label{tab:atrPerLabel2.loo}
\end{table*}

\begin{table*}[h]
	\centering
	\hspace*{-1cm}
	\begin{tabular}{l|c|c|c||c|c|c|c|c|c||c|c|c}
		& $n_e$ & $n_s$ & p99& \textbf{Acc}& \textbf{Gyro}& \textbf{WAcc}& \textbf{Loc}& \textbf{Aud}& \textbf{PS}& \textbf{EF}& \textbf{LFA}& \textbf{LFL} \\
		\hline
		Lying down & 54359 & 47 & 0.38  & 0.62 & 0.59 & 0.71 & 0.54 & 0.69 & 0.76 & 0.81 & 0.79 & \textbf{0.82} \\
		Sitting & 82904 & 50 & 0.49  & 0.58 & 0.58 & 0.68 & 0.58 & 0.62 & 0.71 & \textbf{0.75} & 0.75 & 0.74 \\
		Walking & 11892 & 50 & 0.12  & 0.38 & 0.37 & 0.32 & 0.21 & 0.19 & 0.22 & 0.38 & \textbf{0.39} & 0.39 \\
		Running & 675 & 19 & 0.01  & 0.03 & 0.02 & 0.04 & 0.01 & 0.01 & 0.01 & 0.04 & 0.04 & \textbf{0.04} \\
		Bicycling & 3523 & 22 & 0.04  & 0.19 & 0.15 & 0.23 & 0.16 & 0.12 & 0.15 & \textbf{0.30} & 0.24 & 0.26 \\
		Sleeping & 42920 & 40 & 0.33  & 0.56 & 0.53 & 0.65 & 0.44 & 0.64 & 0.74 & 0.79 & 0.76 & \textbf{0.80} \\
		Lab work & 2898 & 8 & 0.03  & 0.07 & 0.06 & 0.06 & 0.11 & 0.08 & 0.11 & \textbf{0.21} & 0.15 & 0.18 \\
		In class & 2872 & 13 & 0.03  & 0.05 & 0.06 & 0.04 & 0.08 & 0.11 & 0.07 & 0.13 & 0.12 & \textbf{0.14} \\
		In a meeting & 2904 & 34 & 0.03  & 0.06 & 0.04 & 0.05 & 0.06 & 0.11 & 0.07 & \textbf{0.17} & 0.11 & 0.15 \\
		At main workplace & 20382 & 26 & 0.19  & 0.22 & 0.19 & 0.29 & 0.41 & 0.31 & 0.43 & 0.49 & 0.48 & \textbf{0.52} \\
		Indoors & 107944 & 51 & 0.55  & 0.75 & 0.73 & 0.71 & 0.68 & 0.75 & 0.71 & \textbf{0.79} & 0.79 & 0.79 \\
		Outside & 7629 & 36 & 0.08  & 0.21 & 0.20 & 0.18 & 0.16 & 0.16 & 0.20 & 0.23 & \textbf{0.26} & 0.25 \\
		In a car & 3635 & 24 & 0.04  & 0.15 & 0.08 & 0.10 & \textbf{0.27} & 0.13 & 0.16 & 0.23 & 0.22 & 0.23 \\
		On a bus & 1185 & 24 & 0.01  & 0.04 & 0.03 & 0.03 & 0.05 & 0.04 & 0.05 & 0.07 & \textbf{0.07} & 0.07 \\
		Drive (I'm the driver) & 5034 & 24 & 0.06  & 0.21 & 0.09 & 0.16 & \textbf{0.38} & 0.15 & 0.21 & 0.31 & 0.31 & 0.31 \\
		Drive (I'm a passenger) & 1655 & 19 & 0.02  & 0.07 & 0.04 & 0.04 & \textbf{0.15} & 0.07 & 0.08 & 0.13 & 0.12 & 0.11 \\
		At home & 83977 & 50 & 0.49  & 0.67 & 0.65 & 0.65 & 0.63 & 0.71 & 0.69 & 0.75 & 0.76 & \textbf{0.76} \\
		At a restaurant & 1320 & 16 & 0.02  & 0.03 & 0.03 & 0.03 & 0.02 & 0.07 & 0.04 & \textbf{0.11} & 0.07 & 0.10 \\
		Phone in pocket & 15301 & 31 & 0.15  & 0.29 & 0.34 & 0.26 & 0.22 & 0.25 & 0.27 & \textbf{0.38} & 0.37 & 0.37 \\
		Exercise & 5384 & 36 & 0.06  & 0.21 & 0.16 & 0.22 & 0.14 & 0.13 & 0.15 & 0.26 & \textbf{0.26} & 0.24 \\
		Cooking & 2257 & 33 & 0.03  & 0.03 & 0.03 & 0.05 & 0.03 & 0.04 & 0.05 & \textbf{0.08} & 0.07 & 0.07 \\
		Shopping & 896 & 18 & 0.01  & 0.03 & 0.02 & 0.02 & 0.01 & 0.01 & \textbf{0.04} & 0.04 & 0.04 & 0.04 \\
		Strolling & 434 & 8 & 0.01  & 0.01 & 0.02 & 0.01 & 0.01 & 0.01 & 0.02 & 0.02 & \textbf{0.02} & 0.02 \\
		Drinking (alcohol) & 864 & 10 & 0.01  & 0.03 & 0.02 & 0.01 & 0.01 & 0.04 & 0.01 & 0.05 & \textbf{0.06} & 0.05 \\
		Bathing - shower & 1186 & 27 & 0.01  & 0.01 & 0.02 & 0.04 & 0.01 & 0.02 & 0.01 & 0.04 & 0.04 & \textbf{0.05} \\
		\hline
		average & &  & 0.13  & 0.22 & 0.20 & 0.22 & 0.22 & 0.22 & 0.24 & 0.30 & 0.29 & 0.30 \\
	\end{tabular}
	\caption{Leave-one-user-out evaluation performance (F1) of the different classifiers on each label. Part 1 of the labels. For each label $n_e$ is the number of examples and $n_s$ is the number of subjects in the testing (possibly more examples participated in the training). p99 marks the 99$^{th}$ percentile of random scores --- a score above the p99 value has less than 0.01 probability to be achieved randomly. For each label the score of the highest performing classifier is marked in bold.}
\label{tab:f1PerLabel1.loo}
\end{table*}

\begin{table*}[h]
	\centering
	\hspace*{-1cm}
	\begin{tabular}{l|c|c|c||c|c|c|c|c|c||c|c|c}
		& $n_e$ & $n_s$ & p99& \textbf{Acc}& \textbf{Gyro}& \textbf{WAcc}& \textbf{Loc}& \textbf{Aud}& \textbf{PS}& \textbf{EF}& \textbf{LFA}& \textbf{LFL} \\
		\hline
		Cleaning & 1839 & 22 & 0.02  & 0.04 & 0.04 & 0.06 & 0.01 & 0.03 & 0.02 & \textbf{0.07} & 0.06 & 0.06 \\
		Laundry & 473 & 12 & 0.01  & 0.01 & 0.01 & 0.01 & 0.00 & 0.01 & 0.02 & 0.02 & 0.02 & \textbf{0.02} \\
		Washing dishes & 851 & 17 & 0.01  & 0.00 & 0.01 & 0.02 & 0.01 & 0.02 & 0.01 & \textbf{0.03} & 0.03 & 0.03 \\
		Watching TV & 9412 & 28 & 0.10  & 0.14 & 0.11 & 0.12 & 0.12 & 0.19 & 0.17 & 0.23 & 0.22 & \textbf{0.24} \\
		Surfing the internet & 11641 & 28 & 0.12  & 0.14 & 0.16 & 0.16 & 0.14 & 0.17 & 0.15 & \textbf{0.20} & 0.19 & 0.19 \\
		At a party & 404 & 3 & 0.01  & 0.01 & 0.01 & 0.00 & 0.01 & 0.03 & 0.01 & 0.02 & 0.03 & \textbf{0.04} \\
		At a bar & 520 & 4 & 0.01  & 0.01 & 0.01 & 0.01 & 0.01 & 0.02 & \textbf{0.06} & 0.01 & 0.04 & 0.05 \\
		At the beach & 122 & 5 & 0.00  & 0.00 & 0.00 & 0.00 & 0.01 & 0.00 & 0.01 & \textbf{0.05} & 0.02 & 0.02 \\
		Singing & 384 & 6 & 0.00  & 0.01 & 0.01 & 0.00 & 0.01 & 0.01 & 0.01 & 0.00 & \textbf{0.01} & 0.01 \\
		Talking & 18976 & 44 & 0.18  & 0.25 & 0.25 & 0.25 & 0.21 & 0.30 & 0.26 & 0.30 & 0.30 & \textbf{0.30} \\
		Computer work & 23692 & 38 & 0.21  & 0.28 & 0.26 & 0.30 & 0.32 & 0.28 & 0.34 & 0.39 & \textbf{0.39} & 0.39 \\
		Eating & 10169 & 49 & 0.11  & 0.15 & 0.14 & 0.15 & 0.11 & 0.16 & 0.15 & \textbf{0.18} & 0.18 & 0.18 \\
		Toilet & 1646 & 33 & 0.02  & 0.03 & 0.02 & 0.03 & 0.02 & 0.03 & 0.02 & \textbf{0.04} & 0.04 & 0.04 \\
		Grooming & 1847 & 25 & 0.02  & 0.02 & 0.02 & 0.04 & 0.03 & 0.04 & 0.02 & 0.05 & 0.04 & \textbf{0.05} \\
		Dressing & 1308 & 27 & 0.02  & 0.02 & 0.02 & 0.03 & 0.02 & 0.03 & 0.02 & \textbf{0.04} & 0.04 & 0.04 \\
		At the gym & 906 & 6 & 0.01  & 0.01 & 0.01 & 0.02 & 0.01 & 0.02 & 0.03 & \textbf{0.04} & 0.03 & 0.04 \\
		Stairs - going up & 399 & 17 & 0.01  & 0.01 & \textbf{0.02} & 0.01 & 0.01 & 0.01 & 0.00 & 0.01 & 0.02 & 0.01 \\
		Stairs - going down & 390 & 15 & 0.00  & 0.01 & \textbf{0.02} & 0.01 & 0.01 & 0.01 & 0.00 & 0.01 & 0.02 & 0.01 \\
		Elevator & 124 & 8 & 0.00  & \textbf{0.00} & 0.00 & 0.00 & 0.00 & 0.00 & 0.00 & 0.00 & 0.00 & 0.00 \\
		Standing & 22766 & 51 & 0.21  & 0.28 & 0.27 & 0.35 & 0.23 & 0.27 & 0.29 & \textbf{0.35} & 0.34 & 0.35 \\
		At school & 25840 & 39 & 0.23  & 0.30 & 0.30 & 0.29 & 0.42 & 0.37 & 0.40 & \textbf{0.44} & 0.42 & 0.42 \\
		Phone in hand & 8595 & 37 & 0.09  & 0.17 & \textbf{0.17} & 0.11 & 0.13 & 0.12 & 0.12 & 0.16 & 0.17 & 0.16 \\
		Phone in bag & 5589 & 22 & 0.06  & 0.11 & 0.08 & 0.08 & 0.08 & 0.13 & 0.11 & \textbf{0.16} & 0.15 & 0.16 \\
		Phone on table & 70611 & 43 & 0.45  & \textbf{0.59} & 0.58 & 0.51 & 0.48 & 0.51 & 0.56 & 0.55 & 0.59 & 0.58 \\
		With co-workers & 4139 & 17 & 0.05  & 0.06 & 0.06 & 0.07 & 0.07 & 0.11 & 0.09 & 0.13 & 0.12 & \textbf{0.14} \\
		With friends & 12865 & 25 & 0.13  & 0.16 & 0.17 & 0.14 & 0.15 & 0.20 & 0.17 & 0.18 & \textbf{0.20} & 0.20 \\
		\hline
		average & &  & 0.08  & 0.11 & 0.11 & 0.11 & 0.10 & 0.12 & 0.12 & 0.14 & 0.14 & 0.14 \\
	\end{tabular}
	\caption{Leave-one-user-out evaluation performance (F1) of the different classifiers on each label. Part 2 of the labels. For each label $n_e$ is the number of examples and $n_s$ is the number of subjects in the testing (possibly more examples participated in the training). p99 marks the 99$^{th}$ percentile of random scores --- a score above the p99 value has less than 0.01 probability to be achieved randomly. For each label the score of the highest performing classifier is marked in bold.}
\label{tab:f1PerLabel2.loo}
\end{table*}

\end{document}